\documentclass[11pt,letterpaper]{mystyle}
\usepackage{fancyhdr}

\usepackage[utf8]{inputenc} 
\usepackage[T1]{fontenc}
\usepackage[numbers]{natbib}
\usepackage{adjustbox}
\usepackage{breakurl}    
\usepackage{hyperref}
\usepackage[edges]{forest}
\usepackage{subcaption}
\usepackage{soul}
\usepackage{multirow}
\usepackage[utf8]{inputenc} 
\usepackage{booktabs}       
\usepackage{amsfonts}       
\usepackage{nicefrac}      
\usepackage{microtype}     
\usepackage{xcolor}        
\usepackage{colortbl}
\usepackage{enumitem}
\usepackage{amssymb}
\usepackage{wrapfig}
\usepackage{courier}
\usepackage{bxcoloremoji}
\usepackage{CJKutf8}

\newcommand{\github}{\raisebox{-1.5pt}{\includegraphics[height=1.05em]{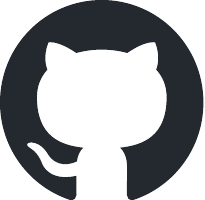}}}

\fancypagestyle{headstyle}{
    \fancyhead[L]{
        \includegraphics[width=80pt]{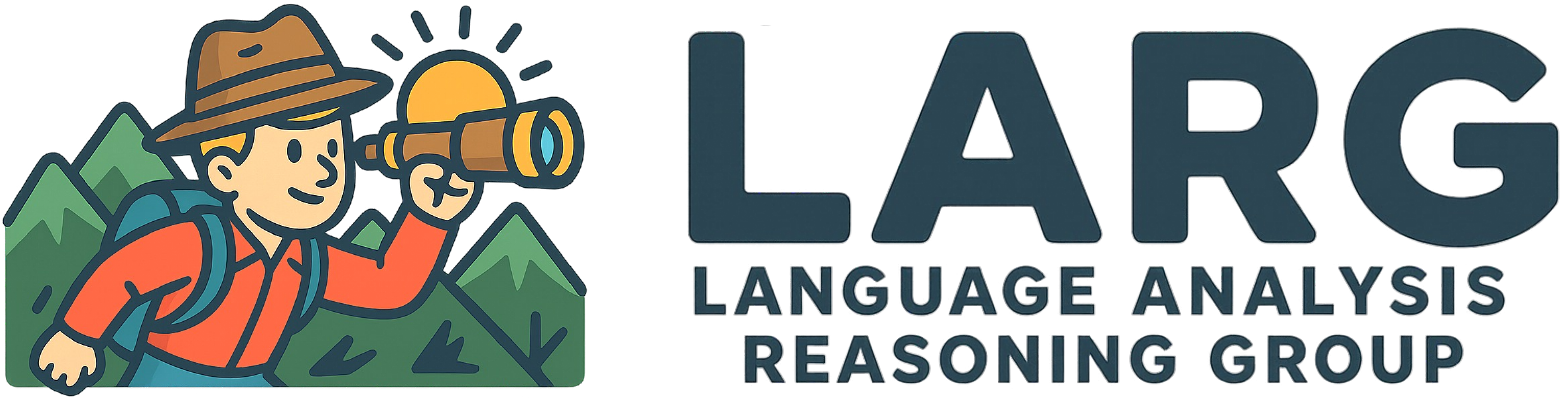}
    }
    \fancyhead[C]{}
    \fancyhead[R]{}

}

\definecolor{new_green}{HTML}{56C596}       
\definecolor{new_sharrow_blue}{HTML}{4DA5D9}
\definecolor{new_blue}{HTML}{3C88C6}  
\definecolor{new_purple}{HTML}{6A0DAD}  
\usepackage{pifont}   
\usepackage{bbding}    
\usepackage{fontawesome} 
\definecolor{my_green}{RGB}{40,154,121}
\definecolor{my_yellow}{RGB}{255,165,0}
\definecolor{my_red}{RGB}{176,46,46}
\newcommand{\correctmark}{\textcolor{my_green}{\ding{52}}} 
\newcommand{\errormark}{\textcolor{my_red}{\ding{56}}}

\definecolor{hidden-red}{RGB}{205, 44, 36}
\definecolor{hidden-blue}{RGB}{194,232,247}
\definecolor{hidden-orange}{RGB}{243,202,120}
\definecolor{hidden-green}{RGB}{34,139,34}
\definecolor{hidden-pink}{RGB}{255,245,247}
\definecolor{hidden-black}{RGB}{20,68,106}
\definecolor{purple}{RGB}{144,153,196}
\definecolor{yellow}{RGB}{255,228,123}
\definecolor{hidden-yellow}{RGB}{255,248,203}
\definecolor{tkcolor}{RGB}{224,223,255}
\definecolor{darkblue}{rgb}{0, 0.40, 0.75}

\hypersetup{colorlinks=true, citecolor=darkblue, linkcolor=darkblue, urlcolor=darkblue}
\tcbset{
  aibox/.style={
    width=\linewidth,
    top=8pt,
    bottom=4pt,
    colback=blue!6!white,
    colframe=black,
    colbacktitle=black,
    enhanced,
    center,
    attach boxed title to top left={yshift=-0.1in,xshift=0.15in},
    boxed title style={boxrule=0pt,colframe=white,},
  }
}

\newtcolorbox{AIBox}[2][]{aibox,title=#2,#1}

\tcbset{
  promptbox/.style={
    width=\linewidth,
    top=8pt,
    bottom=4pt,
    colback=blue!5!white,
    colframe=black,
    colbacktitle=black,
    enhanced,
    center,
    attach boxed title to top left={yshift=-0.1in,xshift=0.15in},
    boxed title style={boxrule=0pt,colframe=white,},
  }
}

\newtcolorbox{PromptBox}[2][]{promptbox,title=#2,#1}

\tcbset{
  takeawaybox/.style={
    width=\linewidth,
    top=8pt,
    bottom=4pt,
    colback=hidden-yellow,
    colframe=black,
    colbacktitle=black,
    enhanced,
    center,
    attach boxed title to top left={yshift=-0.1in,xshift=0.15in},
    boxed title style={boxrule=0pt,colframe=white,},
  }
}

\newtcolorbox{TakeawayBox}[2][]{takeawaybox,title=#2,#1}

\title{OMIBench: Benchmarking Olympiad-Level Multi-Image Reasoning in Large Vision-Language Models}

\author{
Qiguang Chen$^{1,*}$, Chengyu Luan$^{1,*}$, Jiajun Wu$^{2}$, Qiming Yu$^{1}$, Yi Yang$^{2}$, Yizhuo Li$^{1}$, Jingqi Tong$^{3}$, Xiachong Feng$^{4}$, Libo Qin$^{5,6, \coloremojicode{2709}}$, Wanxiang Che$^{1, \coloremojicode{2709}}$\\\vspace{-5pt}
$^{1}$Research Center for Social Computing and Interactive Robotics, Harbin Institute of Technology\\\vspace{-5pt}
$^{2}$Central South University\quad
$^{3}$Fudan University\quad
$^{4}$The University of Hong Kong\\\vspace{-5pt}
$^{5}$Harbin Institute of Technology (Shenzhen)\quad
$^{6}$Text Computing and Cognitive Intelligence Ministry of \\\vspace{-5pt}
Education Engineering Research Center, Guizhou University
}

\begin{abstract}
  \vspace{5mm}
  \textbf{\large Abstract:}
  \vspace{2mm}

Large vision-language models (LVLMs) have made substantial advances in reasoning tasks at the Olympiad level. Nevertheless, current Olympiad-level multimodal reasoning benchmarks for these models often emphasize single-image analysis and fail to exploit contextual information across multiple images. We present OMIBench, a benchmark designed to evaluate Olympiad-level reasoning when the required evidence is distributed over multiple images. It contains problems from biology, chemistry, mathematics, and physics Olympiads, together with manually annotated rationales and evaluation protocols for both exact and semantic answer matching. Across extensive experiments on OMIBench, we observe meaningful performance gaps in existing models. Even the strongest LVLMs, such as Gemini-3-Pro, attain only about 50\% on the benchmark. These results position OMIBench as a focused resources for studying and improving multi-image reasoning in LVLMs.

\vspace{5mm}

$^{*}$ \textit{Equal Contribution}

$^{\coloremojicode{2709}}$ \textit{Corresponding Author}

\vspace{5mm}

\coloremojicode{1F4C5} \textbf{Date}: April 23, 2026

\coloremojicode{1F917} \textbf{Projects}: \href{https://huggingface.co/datasets/LightChen2333/OMIBench}{https://huggingface.co/datasets/LightChen2333/OMIBench}

\github{} \textbf{Code Repository}: \href{https://github.com/LightChen233/OMIBench}{https://github.com/LightChen233/OMIBench}

\coloremojicode{1F4E7} \textbf{Contact}: \href{mailto:qgchen@ir.hit.edu.cn}{qgchen@ir.hit.edu.cn}, \href{mailto:car@ir.hit.edu.cn}{car@ir.hit.edu.cn}, \href{mailto:lbqin@csu.edu.cn}{lbqin@csu.edu.cn}
\end{abstract}

\begin{document}
\maketitle

\vspace{3mm}
\pagestyle{headstyle}
\thispagestyle{empty}

\section{Introduction}

Recent advances in large vision–language models (LVLMs) have enabled strong performance on demanding reasoning tasks, from elementary arithmetic to Olympiad-level problems that require deep domain knowledge and multi-step inference~\citep{lu2022learn,lu2024mathvista,chen-etal-2024-m3cot,wang2025multimodal,liu2025mathematical,he2024olympiadbench}. A central driver of this progress is chain-of-thought (CoT) prompting~\citep{wei2022chain}, which elicits explicit intermediate reasoning steps in natural language~\citep{wang2025multimodal,chen2024unlocking,chen2025towards}. In multimodal settings, these techniques enable LVLMs to fuse visual cues with textual information, yielding substantial gains on single-image Olympiad-level benchmarks~\citep{zhang2024multimodal,cheng2025visual}.

\begin{figure}[t]
    \centering
    \includegraphics[width=\textwidth]{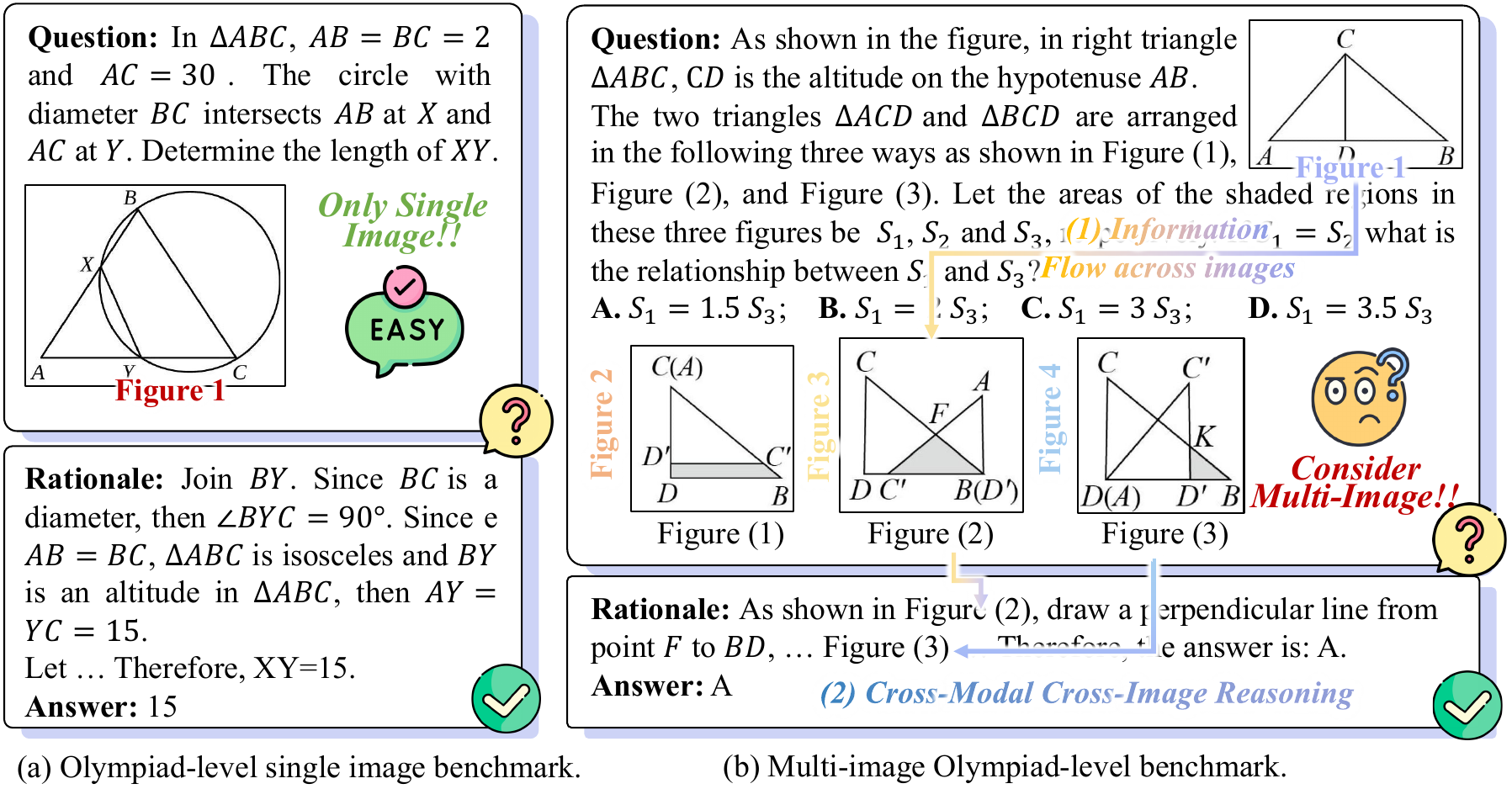} 
    
    \caption{Comparison between existing single-image reasoning benchmarks (OlympiadBench) and our proposed Olympiad-level Multi-Image Reasoning Benchmark (OMIBench).}
    \vspace{-5pt}
    \label{fig:intro}
\end{figure}
\begin{table}[t]
\centering
\scriptsize
\setlength{\tabcolsep}{2.8pt}
\resizebox{\textwidth}{!}{
    \begin{tabular}{lccccccccc}
\toprule
Benchmark & Multi-Image & Maths & Physics & Chemistry & Biology & Difficulty & Rationale & Answer type & Question type \\
\midrule

ScienceQA~\citep{lu2022learn} & \errormark & \correctmark & \correctmark & \correctmark & \correctmark & H & $\thicksim$ 90\% & Text, Num & MC \\ 

MME-CoT~\citep{jiang2025mmecot} & \errormark & \correctmark & \correctmark & \correctmark & \correctmark & H & \correctmark & Text, Num & MC, OE \\
M$^3$CoT~\citep{chen-etal-2024-m3cot} & \errormark & \correctmark & \correctmark & \correctmark & \correctmark & COL & \correctmark & Exp, Text, Num & MC \\
MMReason~\citep{yao2025mmreason} & \errormark & \correctmark & \correctmark & \errormark & \errormark & COL & \correctmark & Exp, Text, Num & OE \\
MathVista~\citep{lu2024mathvista} & \errormark & \correctmark & \errormark & \errormark & \errormark & COL & \errormark & Num & MC, OE \\ 
MathVerse~\citep{zhang2024mathverse} & \errormark & \correctmark & \errormark & \errormark & \errormark & COL & \errormark & Num & MC, OE \\ 
MMMU~\citep{yue2024mmmu} & <10\% & \correctmark & \correctmark & \correctmark & \correctmark & COL & $<18\%$ & Image, Text, Num & MC, OE \\
OlympiadBench~\citep{he2024olympiadbench} & <5\% & \correctmark & \correctmark & \errormark & \errormark & COMP & \correctmark & ALL & ALL \\ 
\midrule
MuirBench~\citep{wang2025muirbench} & \correctmark & \errormark & \errormark & \errormark & \errormark & H & \errormark & Text & MC \\
MMIU~\citep{meng2024mmiu} & \correctmark & \errormark & \errormark & \errormark & \errormark & H & \errormark & Text & MC \\
Blink~\citep{fu2024blink} & \correctmark & \errormark & \errormark & \errormark & \errormark & H & \errormark & Image, Text, Num & MC \\
ReMI~\citep{kazemi2024remi} & \correctmark & \correctmark & \correctmark & \errormark & \errormark & H, COL & \errormark & Image, Text, Num & MC, OE \\

\midrule
OMIBench & \correctmark & \correctmark & \correctmark & \correctmark & \correctmark & COMP & \correctmark & ALL & ALL \\ 
\bottomrule
\end{tabular}
}
\caption{Comparison of representative multimodal benchmarks by image setting, subject coverage, difficulty, rationale, answer types and question types. For \textbf{difficulty}, H: high, COL: college, COMP: competition; For \textbf{answer or choice type}, Num: numeric value, Text: text expression answer or choice, Image: image choices; For \textbf{question type}, MC: multiple-choice, J: judgement, OE: open-ended.}
\vspace{-5pt}
\label{tab:benchmarks}
\end{table}

However, as illustrated in Figure~\ref{fig:intro} (a), existing multimodal Olympiad benchmarks largely remain restricted to single-image question settings~\citep{zhao2024benchmarking,cheng2025evaluating,du2025easy}. In real scientific and technical settings, however, problems often rely on multiple interdependent figures, diagrams, and experimental setups (Figure~\ref{fig:intro}(b))~\citep{alampara2025probing,chen2025ai4research,roberts2024scifibench,liu2024mibench,ji2025mpcc}. Effective multi-image reasoning therefore requires not only interpreting each image, but also (1) \textbf{maintaining a coherent information flow across images}, and (2) \textbf{performing cross-image, cross-modal reasoning} that supports Olympiad-level problem solving.
However, existing benchmarks~\citep{zhang2024mathverse,lu2024mathvista,malkinski2025deep,cheng2025evaluating} only partially capture this multi-image context: they emphasize perception and cross-image reference resolution, but give limited attention to strong semantic and quantitative links across images and modalities in Olympiad-level reasoning. Hence, they offer an incomplete evaluation of multi-image Olympiad-level reasoning, especially in tasks requiring precise interpretation across visuals~\citep{alampara2025probing,cheng2025visual}.

To address this gap, as shown in Table~\ref{tab:benchmarks}, we introduce the Olympiad-level Multi-Image Benchmark (OMIBench), a large-scale suite for evaluating LVLMs’ multi-image information integration and reasoning. OMIBench includes over 1,000 Olympiad-level problems in biology, chemistry, mathematics, and physics, with manually annotated rationales and answers. Each problem contains multiple images that jointly provide the evidence needed for multi-step reasoning and the final solution. OMIBench also offers reasoning-path annotations, enabling fine-grained analyses.

We benchmark representative LVLMs on OMIBench. The results reveal clear limitations, with accuracy below 51\% and drops of up to 15\% relative to single-image settings. Model outputs show recurring failures in visual perception, cross-image association, and cross-modal logical integration; compared with existing multi-image benchmarks, OMIBench produces performance decreases exceeding 20\%.
We also examine several strategies for improving performance, including long chain-of-thought, test-time scaling, ICL, and think-with-image approaches. Long CoT, parallel/sequential scaling, and ICL bring consistent but limited gains, while parameter scaling and think-with-image methods offer little benefit and sometimes reduce performance. These results suggest that progress will likely require more fundamental advances in model architecture and training.

In summary, our contributions are threefold:
\begin{itemize}[leftmargin=16pt, itemsep=0pt, topsep=0pt]
    \item We identify a critical gap in existing literature on evaluating multi-image Olympiad-level reasoning in LVLMs, a setting that requires autonomous cross-image alignment, selection, and integrative reasoning.
    \item We introduce OMIBench, a novel benchmark with over 1,000 Olympiad-level multi-image reasoning tasks spanning chemistry, physics, mathematics, and experimental design. We establish comprehensive baselines by evaluating state-of-the-art LVLMs, exposing major gaps in Olympiad-level multi-image reasoning.
    \item We provide diagnostic analyses and assess diverse enhancement techniques, including long CoT, test-time scaling, in-context learning, and think-with-image methods, to improve LVLM performance on OMIBench and identify promising directions.
\end{itemize}
The dataset and resources are available at \href{https://github.com/LightChen233/OMIBench}{\text{https://github.com/LightChen233/OMIBench}}.

\begin{figure}[t]
    \centering
    \includegraphics[width=0.92\textwidth]{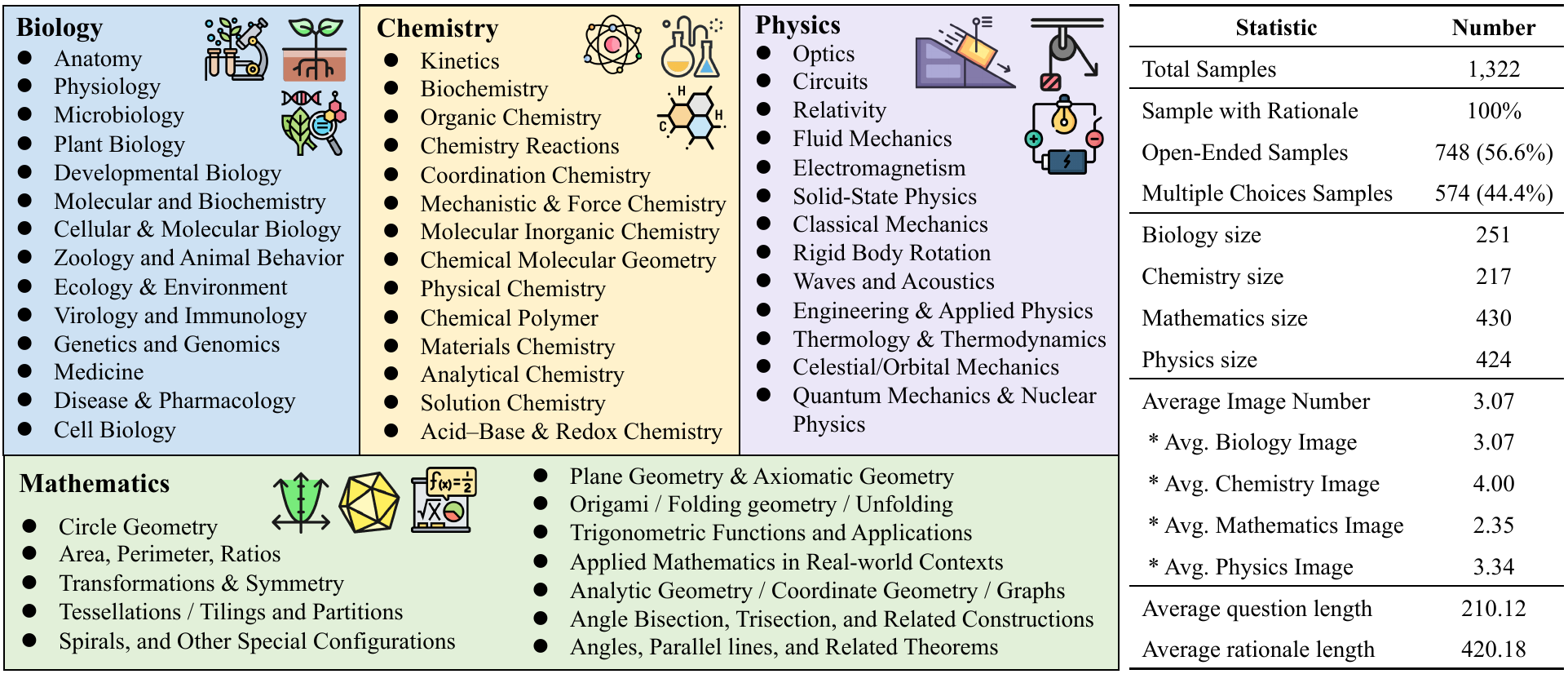} 
    
    \caption{Key statistics of OMIBench, encompassing diverse problem types across Biology, Chemistry, Mathematics, and Physics (over 1.3K samples; average 3.07 images per sample). Images are excluded from token counts.}
    \vspace{-5pt}
    \label{fig:dataset}
\end{figure}

\section{Task Definition}
\label{sec:definition}
Unlike single-image multimodal CoT, multi-image CoT considers a set of images $\mathcal{I} = \{I_1, I_2, \ldots, I_n | n \ge 2\}$, a question $Q$, and a context $C$. The task is to answer $Q$ by integrating evidence across multiple images, where different images may provide complementary information needed for the final answer. Specifically, OMIBench consists of the following two tasks:\vspace{-5pt}

\paragraph{Multiple-Choice Reasoning Task}
Given close set $\mathcal{O} = \{o_1, \dots, o_n\}$ with $n$ options, we first construct a textual prompt $\mathcal{T} = \texttt{Prompt}(Q, C, \mathcal{O})$.
The model then generates a stepwise rationale $\mathcal{R}_m\! =\! \{s_1, \ldots, s_m\}$, with each step $s_i$ defined by:
\begin{equation}
   s_i = \operatorname{argmax}_{s_i\in \mathcal{R}_m} P(s_i|\mathcal{I}, \mathcal{T}).
\end{equation}
Finally, the model selects the final answer $\mathcal{Y}$ from close option set $\mathcal{O}$,
which is denoted as:
\begin{equation}
\mathcal{Y} = \operatorname{argmax}_{o\in \mathcal{O}}
 P (o|\mathcal{R}_m).
\end{equation}

\paragraph{Open-Ended Reasoning Task}
For open-ended problems, we first form an instruction prompt $\mathcal{T} = \texttt{Prompt}(Q, C),$
where $\texttt{Prompt}(\cdot)$ from the question and context, where $\texttt{Prompt}(\cdot)$ denotes the prompting procedure used to format the textual input. Conditioned on $\mathcal{I}$ and $\mathcal{T}$, the model produces a step-by-step rationale $\mathcal{R}_m = \{s_1, \ldots, s_m\}$, with each step generated as:
\begin{equation}
   s_i = \operatorname{argmax}_{s_i\in \mathcal{R}_m} P(s_i|\mathcal{I}, \mathcal{T}).
\end{equation}
Finally, the model arrives at the final answer $\mathcal{Y}$ from open answer space $\mathcal{A}$,
which is denoted as:
\begin{equation}
\mathcal{Y} = \operatorname{argmax}_{\mathcal{A}}
 P (A|\mathcal{R}_m),
\end{equation}
where $\mathcal{Y}$ derives from information in the images and question, requiring the model to integrate visual and textual cues for a coherent answer.

\section{Olympiad-Level
Multi-Image Reasoning Benchmark (OMIBench) }
\label{sec:data_construction}

We build OMIBench to assess whether LVLMs can solve competition-grade scientific problems whose evidence is distributed across multiple images, with coverage across biology, chemistry, mathematics, and physics. Summary statistics are provided in Figure~\ref{fig:dataset} and Table~\ref{tab:dataset_source}. Data construction details can be seen in Appendix~\ref{append:data-construct}.

\subsection{Design Principle}

OMIBench targets the upper bound of Olympiad-level problem solving and supports research on LVLMs for multi-image reasoning in biology, mathematics, chemistry, and physics. Following \citet{he2024olympiadbench}, it reflects the rigor of top competitions. Specifically, OMIBench includes:
\noindent\textbf{Olympiad-level problems:} Biology, mathematics, chemistry, and physics questions from international and national Olympiads for top students, in multiple-choice and open-ended formats, to assess advanced reasoning and intermediate steps.

\noindent\textbf{Expert solutions and rationales:} Each problem includes an expert solution with explicit reasoning. This lowers annotation and evaluation cost, strengthens correctness judgments, and provides supervision for analyzing model reasoning.

\noindent\textbf{Multi-Image reasoning:} Problems that require linking multiple images and their relations, testing cross-image and cross-modal reasoning and integration of visual evidence.

\subsection{Data Annotation}

The data annotation pipeline comprises four phases: data collection \& selection, rationale annotation, quality control, and classification labeling.\vspace{-5pt}

\paragraph{Step 1: Data Collection \& Selection.} OMI-Bench comprises Olympiad-level problems in biology, chemistry, mathematics, and physics, with source distributions summarized in Table~\ref{tab:dataset_source}. The corpus integrates international Olympiads, national and regional contests, and mixed-complexity benchmarks, providing broad Olympiad-level coverage across subfields in these disciplines.

After collecting all PDF files, we use Mathpix OCR to convert problems into Markdown format, and team members manually verify each item to ensure accuracy. The Markdown texts are then normalized into a structured ``Question–Rationale (if available)–Answer'' format. For samples from mixed-complexity benchmarks, expert competitors further select and curate the items. Multilingual questions are translated with Google Translate and subsequently verified by human experts.\vspace{-5pt}

\paragraph{Step 2: Rationale Annotations.}
Most competition datasets omit solution rationales, which are essential for analyzing problem-solving. We therefore build expert-verified rationales via a two-stage pipeline combining LLM-assisted generation (Gemini-2.5-pro-thinking) and human verification.

Specifically, we use two-stage annotation:
(1) LLM generates up to 16 candidate solutions per problem given the reference, retaining those with the correct final answer. If none succeed, we provide the ground-truth answer and regenerate a correct solution, reducing human effort by $\sim$20\%.
(2) Experienced annotators verify and refine the retained rationales by correcting errors, adding missing steps, removing redundancy, and standardizing notation. If a rationale is fundamentally flawed, annotators rewrite it while preserving valid core ideas. A final review ensures correctness, with dataset statistics in Figure~\ref{fig:dataset}.\vspace{-5pt}

\paragraph{Step 3: Quality Control}
To ensure dataset quality, we employ a dual-review protocol in which every problem receives at least one independent audit, complemented by weekly random sampling of 5\% of examples for regression testing on key metrics, error rate, text-image alignment, and solution accuracy. Further, audit feedback is introduced to drive iterative updates to annotation guidelines and targeted retraining, forming a closed-loop quality assurance process that maintains a high-fidelity multimodal competition problem corpus.\vspace{-5pt}

\paragraph{Step 4: Classification Labeling.} OMIBench problems in biology, chemistry, mathematics, and physics fall into two categories: open-ended and multi-choices. The combined Olympiad and high-stakes examination corpus covers a wide range of subfields, as shown in Figure~\ref{fig:dataset}. We first use GPT-4o to generate preliminary topic labels, and then manually assign final topic to ensure consistency and correctness across the corpus.

\begin{table}[t]
\centering
\scriptsize
\setlength{\tabcolsep}{2.4pt}
\resizebox{\textwidth}{!}{
    \begin{tabular}{lcccccccccc}
\toprule
\multirow{2}{*}{\textbf{Model}} & \multicolumn{2}{c}{\textbf{Biology}} & \multicolumn{2}{c}{\textbf{Chemistry}} & \multicolumn{2}{c}{\textbf{Mathematics}} & \multicolumn{2}{c}{\textbf{Physics}} & \multicolumn{2}{c}{\textbf{Total}} \\
\cmidrule{2-11}
 & ACC & Score & ACC & Score & ACC & Score & ACC & Score & ACC & Score \\
\midrule
\rowcolor{gray!10}
\multicolumn{11}{c}{\textit{Instruction LVLMs}} \\
\midrule
InternVL3-1B~\citep{zhu2025internvl3} & 27.41  & 7.97  & 6.54  & 1.84  & 13.39  & 3.26  & 14.86  & 1.42  & 15.40  & 3.33 \\
InternVL3-2B~\citep{zhu2025internvl3} & 31.47  & 20.72  & 12.35  & 11.98  & 17.88  & 7.67  & 14.81  & 9.43  & 18.57  & 11.42 \\
Qwen2.5-VL-3B-Instruct~\citep{bai2025qwen25vl} & 27.89  & 21.12  & 17.05  & 9.68  & 17.91  & 11.16  & 15.57  & 8.25  & 18.91  & 11.87 \\
Qwen2.5-VL-7B-Instruct~\citep{bai2025qwen25vl} & 37.85  & 31.87  & 20.28  & 15.67  & 13.72  & 8.37  & 11.56  & 14.39  & 18.69  & 15.96 \\
InternVL3-8B~\citep{zhu2025internvl3} & 43.57  & 33.07  & 15.80  & 16.13  & 23.30  & 9.30  & 19.53  & 12.74  & 24.71  & 16.04 \\
InternVL3-14B~\citep{zhu2025internvl3} & 47.94  & 38.25  & \cellcolor{new_green!15} 22.66  & \cellcolor{new_purple!15} \textbf{20.74}  & \cellcolor{new_green!15} 24.28  & 11.92  & 21.89  & 16.35  & 27.74  & 19.79 \\
InternVL3-38B~\citep{zhu2025internvl3} & \cellcolor{new_blue!15} 50.92  & 43.03  & 13.36  & \cellcolor{new_green!15} 17.97  & \cellcolor{new_blue!15} 24.35  & 15.81  & \cellcolor{new_purple!15} \textbf{24.81}  & 16.75  & 27.74  & 21.63 \\
InternVL3-78B~\citep{zhu2025internvl3} & 47.41  & \cellcolor{new_green!15} 46.61  & 17.16  & \cellcolor{new_purple!15} \textbf{20.74}  & \cellcolor{new_purple!15} \textbf{27.30}  & \cellcolor{new_green!15} 17.21  & \cellcolor{new_green!15} 22.95  & \cellcolor{new_blue!15} 18.63  & \cellcolor{new_green!15} 28.06  & \cellcolor{new_green!15} 23.83 \\
Qwen2.5-VL-32B-Instruct~\citep{bai2025qwen25vl} & \cellcolor{new_green!15} 48.21  & \cellcolor{new_blue!15} 48.61  & \cellcolor{new_blue!15} 22.93  & \cellcolor{new_blue!15} 19.82  & 24.01  & \cellcolor{new_blue!15} 21.40  & \cellcolor{new_blue!15} 23.54  & \cellcolor{new_purple!15} \textbf{19.81}  & \cellcolor{new_blue!15} 28.28  & \cellcolor{new_blue!15} 25.80 \\
Qwen2.5-VL-72B-Instruct~\citep{bai2025qwen25vl} & \cellcolor{new_purple!15} \textbf{51.79}  & \cellcolor{new_purple!15} \textbf{53.39}  & \cellcolor{new_purple!15} \textbf{23.45}  & \cellcolor{new_blue!15} 19.82  & 24.19  & \cellcolor{new_purple!15} \textbf{27.67}  & 22.47  & \cellcolor{new_green!15} 17.45  & \cellcolor{new_purple!15} \textbf{28.76}  & \cellcolor{new_purple!15} \textbf{27.99} \\
\midrule
\rowcolor{gray!10}
\multicolumn{11}{c}{\textit{Long CoT LVLMs}} \\
\midrule
InternVL3.5-1B~\citep{wang2025internvl35} & 32.46  & 17.93  & 14.29  & 6.91  & 15.12  & 4.65  & 23.58  & 8.02  & 21.56  & 8.62 \\
Qwen3-VL-2B-Instruct~\citep{bai2025qwen3vl} & 27.44  & 19.92  & 12.33  & 5.07  & 11.53  & 6.99  & 12.50  & 8.75  & 14.99  & 9.69 \\
InternVL3.5-2B~\citep{wang2025internvl35} & 34.66  & 25.10  & 16.13  & 13.82  & 20.00  & 11.86  & 22.64  & 12.74  & 23.00  & 14.98 \\
InternVL3.5-30B-A3B~\citep{wang2025internvl35} & 43.03  & 43.43  & 19.35  & 20.28  & 24.42  & 14.19  & 18.87  & 13.21  & 25.34  & 20.43 \\
Qwen3-VL-4B-Instruct~\citep{bai2025qwen3vl} & 43.03  & 36.65  & 17.05  & 11.06  & 27.91  & 23.72  & 18.40  & 13.92  & 25.95  & 20.95 \\
InternVL3.5-14B~\citep{wang2025internvl35} & 51.00  & 40.24  & 19.80  & 21.20  & 27.44  & 19.30  & 18.63  & 20.75  & 27.84  & 24.05 \\
InternVL3.5-241B-A28B~\citep{wang2025internvl35} & \cellcolor{new_green!15} 52.58  & 45.42  & \cellcolor{new_purple!15} \textbf{21.58}  & \cellcolor{new_purple!15} \textbf{25.35}  & 31.16  & 20.47  & 19.58  & \cellcolor{new_green!15} 21.23  & 29.94  & 26.25 \\
InternVL3.5-8B~\citep{wang2025internvl35} & 47.41  & 37.05  & 17.97  & 18.43  & 27.91  & \cellcolor{new_green!15} 32.33  & 17.92  & 17.69  & 26.78  & 26.25 \\
Qwen3-VL-8B-Instruct~\citep{bai2025qwen3vl} & 46.61  & 43.43  & 16.13  & 17.05  & 27.44  & 29.30  & 20.05  & 18.63  & 26.85  & 26.55 \\
InternVL3.5-38B~\citep{wang2025internvl35} & 49.40  & 41.83  & \cellcolor{new_blue!15} 20.28  & \cellcolor{new_green!15} 22.12  & 22.33  & 25.12  & \cellcolor{new_blue!15} 24.53  & \cellcolor{new_blue!15} 23.58  & 27.84  & 27.31 \\
Qwen3-VL-30B-A3B-Instruct~\citep{bai2025qwen3vl} & 48.51  & \cellcolor{new_green!15} 48.61  & 13.29  & 12.90  & \cellcolor{new_green!15} 31.42  & \cellcolor{new_green!15} 32.33  & 20.02  & 20.99  & \cellcolor{new_green!15} 28.03  & \cellcolor{new_green!15} 28.59 \\
Qwen3-VL-235B-A22B-Instruct~\citep{bai2025qwen3vl} & \cellcolor{new_purple!15} \textbf{60.41}  & \cellcolor{new_purple!15} \textbf{63.20}  & \cellcolor{new_green!15} 17.23  & \cellcolor{new_blue!15} 22.58  & \cellcolor{new_blue!15} 37.48  & \cellcolor{new_blue!15} 34.19  & \cellcolor{new_green!15} 23.77  & \cellcolor{new_blue!15} 23.58  & \cellcolor{new_blue!15} 34.11  & \cellcolor{new_blue!15} 34.39 \\
Qwen3-VL-32B-Instruct~\citep{bai2025qwen3vl} & \cellcolor{new_blue!15} 57.62  & \cellcolor{new_blue!15} 58.57  & 14.09  & 20.74  & \cellcolor{new_purple!15} \textbf{44.40}  & \cellcolor{new_purple!15} \textbf{40.70}  & \cellcolor{new_purple!15} \textbf{25.48}  & \cellcolor{new_purple!15} \textbf{25.00}  & \cellcolor{new_purple!15} \textbf{35.87}  & \cellcolor{new_purple!15} \textbf{35.78}  \\
\midrule
\rowcolor{gray!10}
\multicolumn{11}{c}{\textit{Close-sourced LVLMs}} \\
\midrule

GPT-4o-mini~\citep{hurst2024gpt} & 56.57  & 40.24  & 10.81  & 21.66  & 27.49  & 11.86  & 18.25  & 17.22  & 27.31  & 20.58 \\
GPT-4o~\citep{hurst2024gpt} & \cellcolor{new_green!15} 60.96  & 53.00  & 15.67  & 22.58  & 29.69  & 15.49  & 17.10  & 18.92  & 24.05  & 24.88 \\

Gemini-2.5-Flash~\citep{comanici2025gemini} & 58.13  & \cellcolor{new_green!15} 64.54  & 22.42  & 18.43  & 41.16  & 38.37  & 21.21  & 23.35  & 34.91  & 35.25 \\

Gemini-2.5-Pro~\citep{comanici2025gemini} & 59.53  & \cellcolor{new_blue!15} 66.14  & 22.88  & 23.96  & \cellcolor{new_blue!15} 42.28  & 53.49  & \cellcolor{new_green!15} 22.90  & 31.84  & \cellcolor{new_green!15} 36.16  & 44.10 \\
OpenAI-o4-mini~\citep{openai2025o4} & 51.41  & 57.37  & \cellcolor{new_green!15} 23.67  & \cellcolor{new_purple!15} \textbf{32.41}  & \cellcolor{new_green!15} 41.28  & \cellcolor{new_green!15} 56.61  & 19.03  & 35.38  & 33.19  & 45.97 \\
GPT-5-mini~\citep{openai2025gpt5} & 58.96  & 59.36  & 22.12  & 24.42  & 37.44  & \cellcolor{new_blue!15} 56.74  & \cellcolor{new_blue!15} 24.39  & \cellcolor{new_purple!15} \textbf{43.63}  & 34.83  & \cellcolor{new_green!15} 47.73 \\
GPT-5~\citep{openai2025gpt5} & \cellcolor{new_purple!15} \textbf{68.13}  & 62.55  & \cellcolor{new_blue!15} 23.96  & \cellcolor{new_blue!15} {29.03}  & 39.30  & 56.51  & 20.52  & \cellcolor{new_blue!15} 40.80  & \cellcolor{new_blue!15} 36.23  & \cellcolor{new_blue!15} 48.11 \\
Gemini-3-Pro-Preview~\citep{gemini2025gemini3} & \cellcolor{new_blue!15} 64.51  & \cellcolor{new_purple!15} \textbf{71.31}  & \cellcolor{new_purple!15} \textbf{25.52}  & \cellcolor{new_green!15} 25.35  & \cellcolor{new_purple!15} \textbf{55.77}  & \cellcolor{new_purple!15} \textbf{62.56}  & \cellcolor{new_purple!15} \textbf{25.59}  & \cellcolor{new_green!15} 38.92  & \cellcolor{new_purple!15} \textbf{42.79}  & \cellcolor{new_purple!15} \textbf{50.53} \\
\bottomrule
\end{tabular}
}
\caption{Main results on OMIBench, where the \textbf{bold} content denotes the best performance. Here, ``\raisebox{1pt}{\colorbox{new_purple!15}{ \rule[-0.2ex]{0pt}{1.0ex} }}'': best performance, ``\raisebox{1pt}{\colorbox{new_blue!15}{ \rule[-0.2ex]{0pt}{1.0ex} }}'': second performance, ``\raisebox{1pt}{\colorbox{new_green!15}{ \rule[-0.2ex]{0pt}{1.0ex} }}'': third performance. Rows are ordered by total average GPT-Score.}
\vspace{-5pt}
\label{tab:main}
\end{table}

\vspace{-1mm}\section{Main Experiments}\vspace{-1mm}
\label{sec:experiment}
\subsection{Experiments Setup}\vspace{-1mm}
We evaluate advanced open-source and closed-source LVLMs (see Appendix~\ref{append:main-experiment} for additional evaluation details).
Each model generates answers using boxed-format (``$\backslash \text{boxed}\{\cdot\}$'') prompts, and open-source models are deployed on NVIDIA A800 or A100 GPUs.
Temperatures are selected from $[0,1]$.
Model outputs are evaluated using exact-match accuracy and GPTScore, which assesses semantic equivalence under multimodal contextual constraints for open-ended answers (see Appendix~\ref{append:metrics} and Appendix~\ref{append:gptscore-reliability} for more details on the metrics and their reliability).
We report micro-averaged accuracy as the overall metric. 

\vspace{-1mm}\subsection{Main Results}\vspace{-1mm}

Table~\ref{tab:main} presents the overall experimental results, yielding two key findings:\vspace{-5pt}

\paragraph{OMIBench provides a more challenging evaluation framework than existing benchmarks.}
The highest-performing model (Gemini-3-Pro) achieves only 50.53\% on OMIBench, substantially lower than on current benchmarks.
This increased difficulty amplifies performance differences between models, enabling more precise capability comparisons.\vspace{-5pt}

\paragraph{Substantial gaps persist between leading closed- and open-source models, though model scale alone is insufficient.}
Gemini-3-Pro-Preview achieves about 15\% higher accuracy than the best open-source models. However, GPT-4o, despite being closed-source and competitive on complex tasks, achieves accuracy only marginally above open-source models, suggesting that architecture and training strategies beyond parameter count determine performance on challenging benchmarks.

\begin{figure}[t]
    \centering
    \includegraphics[width=0.90\textwidth]{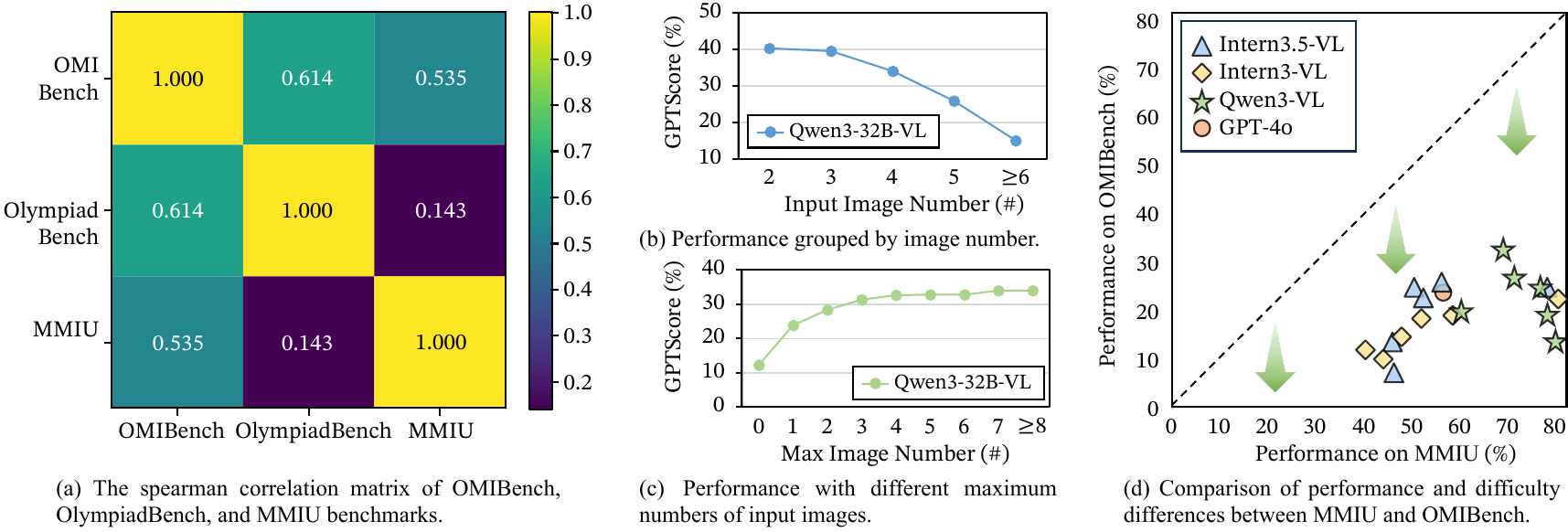} 
    \caption{The performance analysis for benchmark feature analysis and statistics.}
    \vspace{-5pt}
    \label{fig:statistics}
\end{figure}

\begin{figure}[t]
    \centering
    \includegraphics[width=0.88\textwidth]{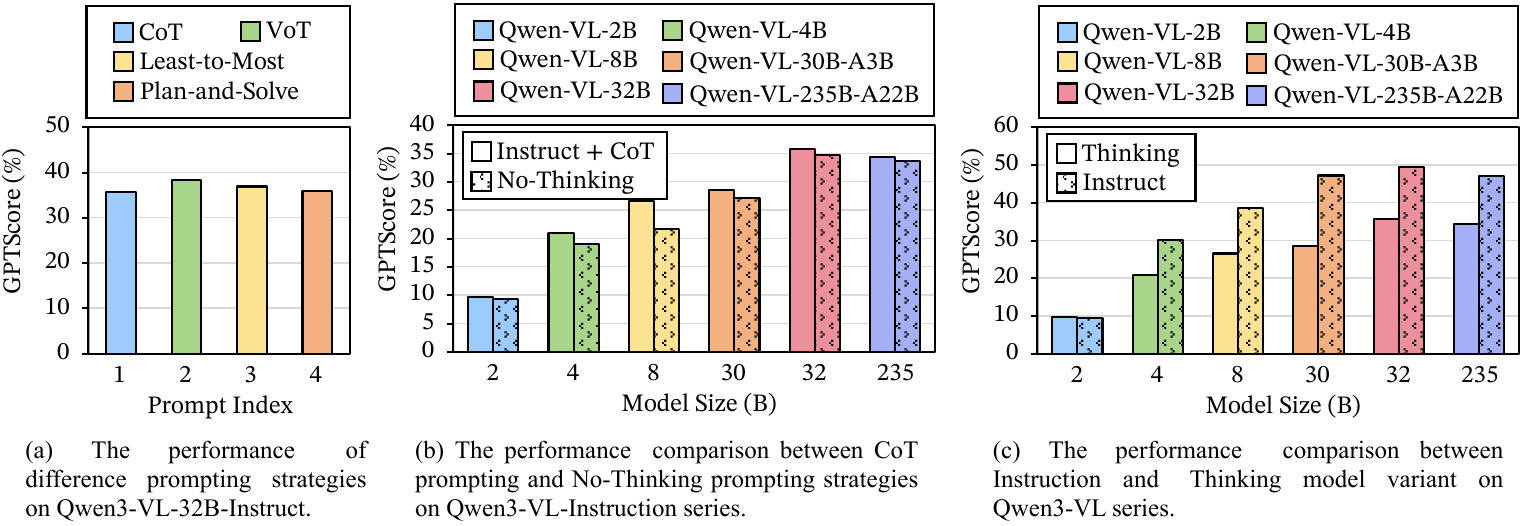} 
    \caption{Performance analysis on Long CoT Strategies. More details can be seen in Appendix~\ref{app:long-cot-details}.}
    \vspace{-5pt}
    \label{fig:thinking}
\end{figure}

\vspace{-1mm}\section{What's essential in OMIBench?}\vspace{-1mm}
We analyze OMIBench in the context of existing multi-image benchmarks and isolate what makes its Olympiad-style multi-image reasoning distinct. More details are shown in Appendix~\ref{append:olympiad-thinking-requirement} \& \ref{append:multi-image-requirement} \& \ref{append:combined-requirement}.\vspace{-5pt}

\paragraph{OMIBench needs deeper Olympiad-level cross-modal reasoning.}
To gauge Olympiad-level reasoning demands, we compare OMIBench with single-image OlympiadBench using the same LVLMs. Figure~\ref{fig:statistics}(a) shows only moderate Spearman correlation across models ($\rho=0.614 < 0.7$), suggesting that multi-image inputs shift relative rankings even on similar Olympiad problems. Accordingly, Gemini-3.0-Pro drops from 75.67\% accuracy on OlympiadBench to 50.53\% on OMIBench (>25\% absolute), indicating the added difficulty of multi-image Olympiad reasoning.
To further probe these demands, we sample 100 problems and rate rationales from o4-mini and Gemini-3.0-Pro. The human review finds logical errors in 46\% of key steps, exposing a gap between fluent rationales and correct reasoning. This gap calls for stronger rationale generation to reach Olympiad-level reasoning depth.\vspace{-5pt}

\paragraph{OMIBench needs stronger awareness of the information flow across images.}
Generally, MMIU~\citep{meng2024mmiu} targets basic multi-image understanding, whereas OMIBench demands complex reasoning across images. To confirm this, we compare model performance. As shown in Figure~\ref{fig:statistics} (a), OMIBench shows moderate Spearman correlation with MMIU (< 0.7): it links to multi-image tasks but Olympiad-level difficulty alters model rankings.
Beyond this, we further examine multi-image information integration. As shown in Figure~\ref{fig:statistics} (b), single-image accuracy reaches 40\%, dropping below 15\% for inputs with $\ge 6$ images. Restricting instances to one image (Figure~\ref{fig:statistics} (c)) causes at least a 10\% absolute performance drop versus using all images, revealing LVLMs' struggles with cross-image integration.\vspace{-5pt}

\paragraph{OMIBench needs combined Olympiad-level cross-image and cross-modal reasoning.}
As shown in Figure~\ref{fig:statistics} (a), MMIU and OlympiadBench yield poorly aligned model rankings, whereas OMIBench is more consistent with both, suggesting that it more faithfully captures the joint demands of multi-image and Olympiad-style problems. The performance comparison in Figure~\ref{fig:statistics} (d) further illustrates this distinction: MMIU primarily evaluates basic visual understanding (even the weakest models achieve >40\% accuracy), whereas OMIBench requires substantially deeper reasoning (the best model reaches only around 40\%). This pronounced performance gap highlights the increased difficulty of OMIBench and its value for stress-testing LVLMs on both multi-image integration and Olympiad-level reasoning.\vspace{-5pt}

\paragraph{Mistake Analysis.}
We further analyze GPTScore-annotated incorrect samples to identify common failure modes, grouping them into three categories.
As shown in Figure~\ref{fig:error}, visual perception failures account for 13\% of errors, cross-image association failures for 29\%, and logical reasoning fallacies for 41\%. These distributions reveal persistent challenges in complex visual interpretation, multi-image integration, and logical consistency, highlighting the need for targeted advances to improve LVLM performance.\vspace{-5pt}

\paragraph{Cross-image reasoning remains the main bottleneck.}
To disentangle cross-image reasoning from confounders such as increased visual information, longer inputs, or OCR noise, we construct an \emph{information-equivalent single-image} control by concatenating all images for each problem into a single composite image while keeping the question text and answer choices unchanged. This preserves the total visual and textual information, logical difficulty, input length, and OCR-related noise. Detailed results are provided in Appendix~\ref{append:single-vs-multi}. The information-equivalent single-image setting consistently outperforms the original multi-image setting, indicating that the multi-image \emph{organization} itself, which requires models to autonomously align, filter, and integrate evidence across images, is the primary source of the observed performance gap rather than visual volume or OCR noise.

\paragraph{Human-expert baselines confirm the difficulty of OMIBench.}
To calibrate the absolute difficulty of OMIBench against human performance, we conduct an initial human-baseline study on a 52-problem subset. Human experts achieve above $80\%$ accuracy, and trained non-experts exceed $57\%$, while the strongest current model (Gemini-3-Pro) reaches only $48.08\%$, leaving a gap of more than 30 points to experts and around 10 points even to trained non-experts. Additional results are provided in Appendix~\ref{append:human-baseline}.

\section{How to improve on OMIBench?}

\begin{figure}[t]
    \centering
    \includegraphics[width=0.92\textwidth]{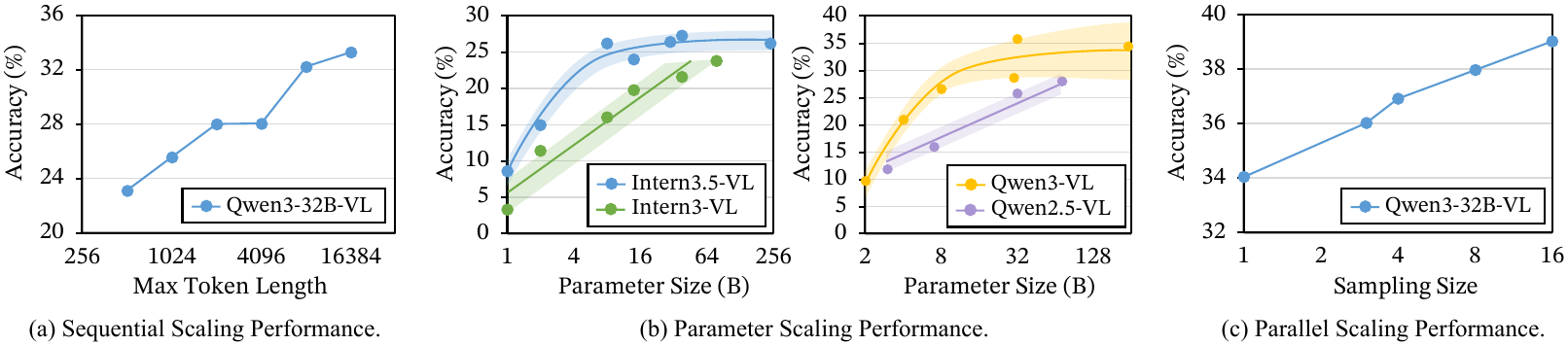} 
    
    \caption{Performance analysis on 3-dimensional Test-Time Scaling paradigms. See Appendix~\ref{append:tts-exp} for more details.}

    \vspace{-5pt}
    \label{fig:scaling}
\end{figure}

\begin{figure}
    \centering
    \includegraphics[width=0.88\textwidth]{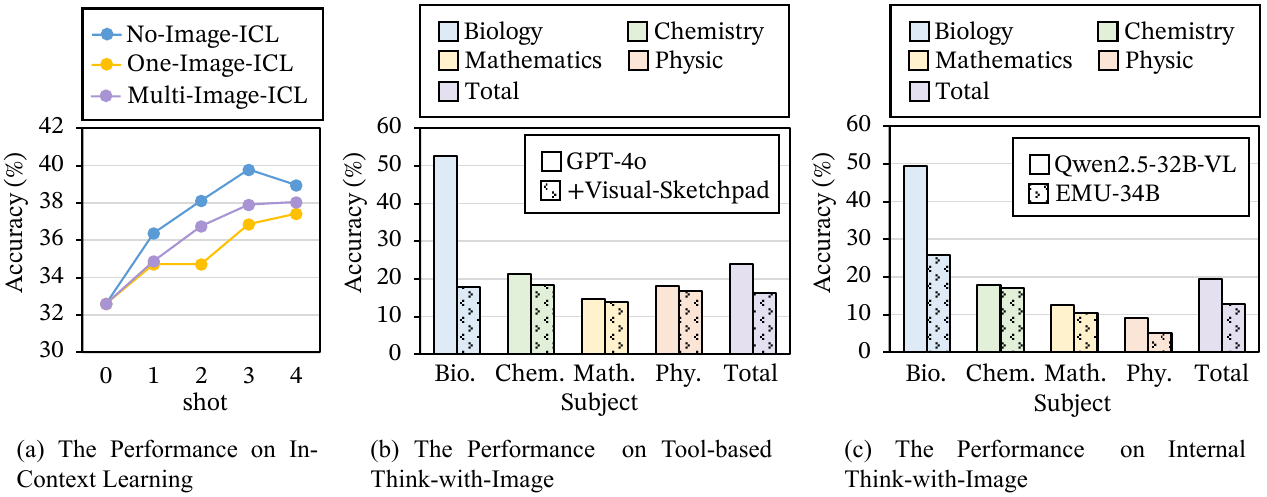} 
    
    \caption{The performance analysis on multimodal In-Context Learning and Thinking-with-Image paradigms.}
    \vspace{-5pt}
    \label{fig:think-with-image}
\end{figure}

\subsection{Can Long CoT Strategies Help?}
\label{sec:long-cot}
\paragraph{Prompting methods that usually work do not significantly improve performance on OMIBench.} Prior work has found that chain-of-thought (CoT) prompting can improve model performance on Olympiad-level reasoning tasks~\citep{chen2025towards,li2025system}. We test whether these gains transfer to OMIBench by comparing widely used CoT prompting strategies, asking whether prompt engineering can narrow the gap or whether improvements remain marginal. As shown in Figure~\ref{fig:thinking} (a), we observe no statistically significant differences across prompts, indicating that current CoT prompting yields only limited gains on OMIBench.\vspace{-5pt}

\paragraph{Long CoT thinking model variants significantly improve performance on OMIBench compared with prior benchmarks.} We further evaluate a “no-thinking” prompting setting and find that both reasoning-oriented and non-reasoning LVLMs still struggle on OMIBench. As shown in Figure~\ref{fig:thinking} (b), enabling or disabling explicit “thinking” yields little performance change. We then compare the “thinking” and “instruct” variants of Qwen3-VL to assess whether more advanced Long CoT paradigms improve multimodal reasoning. As shown in Figure~\ref{fig:thinking} (c), the thinking variant outperforms the instruct variant on OMIBench by about 10\%, substantially improving Olympiad-level performance. This gain is notably larger than that on earlier single-image reasoning benchmarks such as MathVista (<5\%)~\citep{bai2025qwen3vl}.

\subsection{Can Test-Time Scaling Help?}\vspace{-1mm}

To assess test-time scaling on OMIBench, we vary three orthogonal axes (parallel sampling, sequential reasoning depth, and model size), and obtain the following observations:\vspace{-5pt}

\paragraph{Sequential scaling remains effective in OMIBench.}
We test whether longer test-time reasoning improves performance by varying the maximum reasoning length. The resulting accuracy–reasoning-length curve characterizes the marginal returns (and potential saturation) of additional sequential inference computation. As shown in Figure~\ref{fig:scaling}(a), increasing the token budget from 512 to 16{,}384 yields a near-monotonic accuracy gain, indicating that such scaling remains effective on OMIBench, where MMIU and OlympiadBench performance becomes comparable.
\vspace{-5pt}

\paragraph{Parameter scaling limits on OMIBench necessitate increased activated rather than total activable parameters.}
We evaluate test-time scaling versus parameter count using models from 1B to 235B parameters under identical inference. Unlike OlympiadBench and MMIU (Figure~\ref{fig:scaling}(b)), InternVL and QwenVL plateau on OMIBench at $\sim$25\% and $\sim$35\% GPTScore, respectively. This plateau is mainly associated with Mixture-of-Experts (MoE) models, where added capacity may not be activated at inference. In contrast, performance plotted against activated parameters shows a positive, near-linear trend, implying that multi-image Olympiad tasks require more concurrently active parameters rather than a larger inactive pool.\vspace{-5pt}

\paragraph{Parallel scaling still works in OMIBench.}  
With temperature set to 0.6 and self-consistency applied, accuracy in Figure~\ref{fig:scaling}(c) improves monotonically as $k$ increases and exhibits approximately log-linear scaling in the number of samples. These results indicate that parallel scaling remains effective on OMIBench.

\subsection{Can In-Context Learning Help?}\vspace{-1mm}

Inspired by Multimodal In-Context Learning (MM-ICL)~\citep{qin2024factors}, we ask whether curated in-context examples can improve multimodal reasoning without parameter updates.\vspace{-5pt}

\paragraph{MM-ICL offers a limited logical connection of multi-image context.}
Figure~\ref{fig:think-with-image}(a) shows that multi-image ICL exceeds single-image ICL, yet both underperform No-Image-ICL, indicating that current LVLMs benefit more from textual context than from multi-image visual context for cross-image reasoning.\vspace{-5pt}

\paragraph{Multiple visual logical connections remain inferior to textual connections.}
This differs from prior single-image findings~\citep{chen-etal-2024-m3cot,qin2024factors}, suggesting that MM-ICL can link multimodal context to some extent but remains weaker than text-based connection.

\subsubsection{Can Thinking with Images Help?}

Further, we examine whether models can effectively \emph{think with images (TwI)} by generating or manipulating intermediate visual artifacts during reasoning, rather than producing only textual rationales~\citep{cheng2025comt,cheng2025visual,su2025thinking}. Additional details are provided in Appendix~\ref{append:think-with-image}.\vspace{-5pt}

\paragraph{Current tool-based TwI works in single image domain but fails in OMIBench.}
Tool-based TwI with GPT-4o and VisualSketchpad still suffers substantial performance degradation on OMIBench in Figure~\ref{fig:think-with-image} (b), indicating limited transfer from single-image tasks.\vspace{-5pt}

\paragraph{Current internal TwI works in single image domain but fails in OMIBench.}
Internal TwI with Emu-3.5-34B likewise incurs large performance degradation in Figure~\ref{fig:think-with-image} (c) and even falls below Qwen-2.5-32B-VL, further underscoring its limited transfer from single-image tasks.\vspace{-2mm}

\subsection{Can SFT or  Tools Close the Gap?}
\label{sec:sft-tools}
Beyond prompting and test-time strategies, we ask whether (1) supervised fine-tuning (SFT) on existing multi-image data, or (2) integration of external visual tools, can substantially narrow the gap on OMIBench. Detailed numbers are reported in Appendix~\ref{append:sft-tools} (Tables~\ref{tab:sft} and~\ref{tab:tools}).\vspace{-5pt}

\paragraph{SFT on existing multi-image data yields only limited gains.}
We fine-tune InternVL3.5-8B and Qwen3-VL-8B-Instruct on two representative multi-image SFT datasets: \textbf{CMMCoT}~\citep{cmmcot2026} and \textbf{MMDU}~\citep{mmdu2024}. Naive SFT on the simpler MMDU even slightly degrades performance ($\sim$0.5--1\%), while SFT on the more reasoning-intensive CMMCoT consistently improves both backbones, but with average gains still below $\sim$1.5--2\%. This shows that the field currently lacks training data tailored to \emph{Olympiad-level} multi-image reasoning, and that closing the gap on OMIBench likely requires fundamentally new training resources.\vspace{-5pt}

\paragraph{External visual tools help only when paired with strong base models.}
We evaluate three external-tool integration frameworks, Visual Sketchpad~\citep{hu2024visualsketchpad}, SlowPerception~\citep{wei2024slowperception}, and CogFlow~\citep{cogflow2026}, on top of GPT-4o and GPT-5. When the base model is weaker (GPT-4o), tool augmentation generally \emph{degrades} performance, as the model frequently fails to invoke visual tools correctly in multi-image scenarios. With a stronger backbone (GPT-5), tool augmentation yields modest additional gains in some subjects, but the overall improvement remains limited, indicating that reliably solving OMIBench requires stronger underlying intelligence to orchestrate tool use, rather than tools alone.

\vspace{-1mm}\section{Related work}\vspace{-1mm}

\paragraph{Competition Benchmarks.}
Several benchmarks assess LLM reasoning using competition-style questions. MATH~\citep{hendrycks2021measuring} and OlymMATH~\citep{sun2025challenging} collect high-school mathematics contest problems across diverse topics and require multi-step reasoning. AGIEval~\citep{zhong2024agieval} covers multiple subjects, including mathematics and physics, with questions drawn from competitive examinations. OlympiadBench~\citep{he2024olympiadbench} targets Olympiad-level mathematics and physics and provides per-problem rationales to enable deeper analysis of model reasoning. In addition, AIME~\citep{aime2024,aime2025} and AMC~\citep{amc2023} use authentic contest problems that probe a broad range of mathematical skills and concepts.\vspace{-5pt}

\paragraph{Multimodal Benchmarks.}
To evaluate multimodal geometric reasoning, datasets such as Geometry3K~\citep{lu2021inter}, GeoQA~\citep{chen2021geoqa}, GeoQA+~\citep{cao-xiao-2022-augmented}, and UniGeo~\citep{chen2022unigeo} pair natural-language problems with diagrams~\citep{ji2025mpcc}. MathVista~\citep{lu2024mathvista} and MathVerse~\citep{zhang2024mathverse} assess broader multimodal mathematical reasoning in vision-language models. ScienceQA~\citep{lu2022learn}, MMMU~\citep{yue2024mmmu}, and M3CoT~\citep{chen-etal-2024-m3cot} extend evaluation across multiple disciplines~\citep{jiang2025mmecot}. OlympiadBench~\citep{he2024olympiadbench} targets advanced scientific reasoning through Olympiad problems in mathematics and physics, while Physics Big~\citep{timur2024physics} provides a large-scale collection of physics competition problems for evaluating quantitative problem solving.

Recent work has also examined multi-image understanding in LVLMs. Mementos~\citep{wang2024mementos} studies narrative and temporal reasoning over image sequences; MC-Bench~\citep{xu2025mcbench} evaluates multi-context visual grounding; and MANTIS~\citep{jiang2024mantis} offers interleaved vision--language instruction-tuning data with broad, general-domain difficulty. Although these benchmarks effectively test cross-image reference, alignment, and commonsense composition, they mainly emphasize narrative or perceptual integration rather than the tight semantic and quantitative coupling required by competition-level scientific problems.

Overall, existing benchmarks have advanced the evaluation of multimodal reasoning, but they still focus largely on single-image settings or relatively shallow multi-image perception, and seldom capture Olympiad-level multi-step reasoning. OMIBench addresses this gap by combining Olympiad-level scientific reasoning with evidence distributed across multiple interdependent images, requiring coherent image--image and image--text reasoning to derive the final answer.

\vspace{-1mm}\section{Conclusion}\vspace{-1mm}
This work introduced OMIBench, a large-scale multi-image Olympiad-level benchmark for evaluating LVLMs on complex multi-image reasoning. Experiments show substantial performance drops relative to single-image tasks, driven by failures in multi-image integration and grounded cross-modal reasoning. These findings establish multi-image reasoning as a central challenge and motivate advances beyond prompting.

\section*{Ethical Considerations}
In this paper, we introduce OMIBench, a demanding multimodal benchmark for assessing mathematical and physical reasoning in current large models and future AGI systems. We outline the dataset construction pipeline, encompassing data collection from official sources only, OCR processing, cleaning, deduplication, and expert annotations.

Each problem includes rigorous annotations, with an evaluation script provided for reproducible model assessment. OMIBench thus supports advances in AI scientific reasoning. To ensure reproducibility and curb carbon-intensive redundant computation, we will release the dataset and scripts publicly. All experiments adhere to relevant model and data licenses.

\section*{Limitations}
OMIBench still has several limitations in evaluation. First, some questions require open-ended textual reasoning, such as multi-part solutions or responses containing multiple valid scientific statements, and therefore cannot yet be evaluated fully reliably using symbolic tools such as SymPy. These cases still require model-based or human review. Second, even by GPTScore evaluation, it may still under-credit creative solutions or partially correct answers in open-ended settings.
A further limitation concerns dataset construction. Due to the complexity and resource requirements of building a multimodal scientific reasoning benchmark, although OMIBench covers multiple disciplines and problem formats, it does not yet cover the full range of multi-image Olympiad-style reasoning found in real educational and scientific settings.

\section*{Acknowledgements}
We gratefully acknowledge the support of the National Natural Science Foundation of China (NSFC) via grant 62236004, 62476073, 92570120 and 62306342. This work was supported by the Scientific Research Fund of Hunan Provincial Education Department (24B0001). This work
was sponsored by the Excellent Young Scientists Fund in Hunan Province (2024JJ4070), the Science and Technology Innovation Program of Hunan Province under Grant 2024RC3024. This study was also funded by the Open Project of the Text Computing and Cognitive Intelligence Ministry of Education Engineering Research Center (No. TCCI250101).

\bibliographystyle{./refstyle}
\bibliography{ref}
\appendix
\newpage
\begin{center}
\textbf{\LARGE Appendix}
\end{center}

\vspace{-2mm}\section{Data Construction Details}\vspace{-1mm}
\label{append:data-construct}
OMIBench was constructed via a rigorous multi-stage pipeline that aggregates high-quality, reasoning-intensive problems from diverse global sources. Key steps are outlined below.

\vspace{-2mm}\subsection{Details of Data Collection}\vspace{-1mm}
Our collection strategy prioritized three criteria: (1) difficulty, requiring multi-step chain-of-thought reasoning~\citep{chen2026molecular}; (2) diversity, spanning text-only and multimodal contexts; and (3) authority, drawing on established competitions and vetted academic benchmarks.

The data acquisition process followed two primary streams: (1) \textbf{Original contest archival.} We manually collected official problem sets from international and national Olympiad archives (e.g., ICHO, IMO, CPHO) and regional tournaments (e.g., ASO, CUPT). For sources available only as PDFs, we used optical character recognition (OCR) tools specialized for scientific notation (e.g., Mathpix) to extract textual content and \LaTeX{} equations, while high-resolution figures were cropped and preserved for multimodal evaluation.
(2) \textbf{Integration of existing benchmarks.} To broaden coverage, we adapted images and questions from recent open-source benchmarks, including OlympiadBench~\citep{he2024olympiadbench}, Mv-MATH~\citep{wang2025mv}, and EMMA~\citep{hao2025can}. For these sources, we standardized the data format to unify the representation of questions, images, and ground-truth answers across disciplines.

Table~\ref{tab:dataset_source} lists the sources for each discipline and the rationale for their inclusion.\vspace{-5pt}
\paragraph{Biology:} Data were sourced from the \textit{Australian Science Olympiads (ASO)} and the \textit{Indian National Biology Olympiad (INBO)}. These contests feature long-context problem statements that require synthesizing facts with experimental data, testing LLMs' capacity for evidence-based reasoning.\vspace{-5pt}
\paragraph{Chemistry:} This subset combines classical analytical problems from the \textit{International Chemistry Olympiad (ICHO)} with interdisciplinary challenges. We include questions from the \textit{British Biology Olympiad (BBO)} that overlap with biochemistry, alongside the \textit{Chemistry Race} and  modified \textit{EMMA} instances, to diversify problem formats from organic synthesis planning to physical chemistry calculations.\vspace{-5pt}
\paragraph{Mathematics:} To capture high-level symbolic reasoning, we aggregate problems from the \textit{IMO}, \textit{CMO}, and \textit{EGMO}. Supplemented by \textit{Mv-MATH} and \textit{OlympiadBench}, this component is heavily multimodal, with plane geometry diagrams and function graphs that require aligning visual perception with symbolic deduction.\vspace{-5pt}
\paragraph{Physics:} We incorporate problems from the \textit{China Undergraduate Physics Tournament (CUPT)} and the \textit{Chinese Physics Olympiad (CPHO)}, selected for their emphasis on physical intuition and complex modeling of real-world phenomena. Data from \textit{Physics-Big} provide additional coverage of mechanics and electromagnetism.

\begin{table}[t]
\centering
\footnotesize
\setlength{\tabcolsep}{4pt}
\resizebox{0.54\textwidth}{!}{
    \begin{tabular}{ll}
\toprule
\textbf{Suject} & \textbf{Source} \\
\midrule
\multirow{2}{*}{Biology} & Australian Science Olympiads (ASO)~\footnote{https://www.asi.edu.au/program/australian-science-olympiads/}\\
 &Indian National Biology Olympiad (INBO)~\footnote{https://www.ibo-info.org/en/} \\
 \midrule
\multirow{4}{*}{Chemistry} & British Biology Olympiad (BBO)~\footnote{https://ukbiologycompetitions.org/british-biology-olympiad/}\\
 & International Chemistry Olympiad (ICHO)~\footnote{https://www.ichosc.org/}\\
 & Chemistry Race~\citep{hrubes2021chemistry}\\
 & EMMA~\citep{hao2025can} \\
 \midrule
\multirow{5}{*}{Mathematics} & International Mathematical Olympiad (IMO)~\footnote{https://www.imo-official.org/}\\
 & Chinese Mathematical Olympiad (CMO)~\footnote{https://sgmathsociety.org/simo/china-mathematical-olympiad-cmo/}\\
 & European Girls' Mathematical Olympiad (EGMO)~\footnote{https://www.egmo.org/}\\
 & OlympiadBench~\citep{he2024olympiadbench} \\
 & Mv-MATH~\citep{wang2025mv} \\
 \midrule
\multirow{4}{*}{Physics} & China Undergraduate Physics Tournament (CUPT)~\footnote{https://www.cupt-iypt.com/}\\
 & Chinese Physics Olympiad (CPHO)~\footnote{https://cpho.pku.edu.cn/}\\
 & OlympiadBench~\citep{he2024olympiadbench} \\
 & Physics-Big~\citep{timur2024physics} \\
\bottomrule
    \end{tabular}
}
\caption{Source statistics of the constructed OMIBench dataset.}
\label{tab:dataset_source}
\end{table}

\vspace{-2mm}\subsection{Details of Format Conversion and Data Selection.}\vspace{-1mm}
To further ensure dataset quality, we apply additional filtering and translation checks beyond the main pipeline described in the paper.\vspace{-5pt}
\paragraph{Format Conversion.} First, we convert all PDF files to Markdown using Mathpix OCR and normalize them to a ``Question–Rationale (if available)–Answer'' schema.\vspace{-5pt}

\paragraph{Format Filtering.} Team members review each item to confirm that (i) the problem statement is complete and legible, and (ii) the answer is well defined and unambiguous. Items with severe OCR errors, missing essential information (e.g., truncated questions or missing answers), duplicated content, or unresolved formatting issues (such as unreadable formulas or diagrams) are removed from the final set.\vspace{-5pt}

\paragraph{Difficulty Curation.} For benchmarks spanning multiple difficulty levels, experienced competition participants further curate the pool by excluding trivial, overly domain-specific, or stylistically inconsistent problems, so that the remaining items better align with the intended reasoning skills and difficulty.\vspace{-5pt}

\paragraph{Multilingual Translation and Verification.} For multilingual components, all non-English questions are first translated into English using Google Translate. Human annotators fluent in English then verify and, when necessary, correct the translations to preserve the original semantics, mathematical notation, and any subtle constraints or assumptions. During this process, annotators flag and resolve ambiguities (e.g., multiple plausible interpretations of a term or condition), and any items whose meaning cannot be reliably disambiguated are discarded.

\vspace{-2mm}\subsection{Details of Rationale Annotations}\vspace{-1mm}

Most existing competition-style datasets provide only final answers or brief solution sketches, which are insufficient for analysing model reasoning behaviours.
To address this limitation, we construct expert-verified, step-by-step rationales for each problem using a two-stage pipeline that combines LLM generation with careful human verification.

\vspace{-2mm}\subsubsection{LLM-Assisted Rationale Generation}\vspace{-1mm}
We first use Gemini-2.5-pro-thinking to generate multiple candidate solutions for each problem.\vspace{-5pt}

\paragraph{Model Prompting:}
For every problem, the input to the model includes: the problem statement; all auxiliary information required to solve the problem (e.g., provided figures, tables, or input–output formats), when available.
Concretely, we use a prompt template that instructs the model to ``Let's think step-by-step!'' and enforces a standardised answer format.
Each candidate must include both an explicit reasoning trace and an explicit final answer.
Specifically, the model prompt is structured as follows:
\begin{PromptBox}{{LLM Rationale Generation Prompt-P1}}
{You are an expert problem solver. Given the following problem statement and any auxiliary information, provide a clear step-by-step solution that leads to a final answer. Explain each step of your reasoning, and format your response as: \{Reasoning: ... Final Answer: ...\}}.\\

\textbf{Problem Statement:} \texttt{[Insert problem statement here]}\\
\textbf{Auxiliary Information:} \texttt{[Insert figures, tables, or other relevant data here]}
\end{PromptBox}
\noindent For each problem, Gemini-2.5-pro-thinking samples up to \(16\) candidate solutions at moderate temperature (\(0.6\)) to balance diversity and coherence.\vspace{-5pt}

\paragraph{Rationale Filtering:}
After generation, we automatically filter candidates, retaining only those whose final answer matches the known correct answer under the official evaluation protocol. If at least one such candidate exists, all matching candidates are kept as potential rationales. If none of the \(16\) candidates is correct, we append the corresponding reference solution sketch or official final answer to the prompt as ``\textbf{Reference Solution:} \texttt{[Insert reference solution here]}'' to trigger generation of a correct rationale.
To avoid trivial rationales that merely restate the answer, we apply simple automatic heuristics to discard degenerate candidates, such as explanations shorter than 50 tokens or responses that only paraphrase the question or the reference answer without intermediate steps.

Overall, this LLM-assisted stage reduces human annotation effort by about \(60\%\) relative to fully manual authoring, while still providing a rich pool of candidate rationales for each problem.

\vspace{-2mm}\subsubsection{Human Verification and Refinement}\vspace{-1mm}
Based on the filtered candidate rationales, experienced annotators are required to verify and refine all LLM-generated rationales.
Annotators are graduate-level students or domain experts with strong backgrounds in mathematics, computer science, or related fields, and receive detailed written guidelines and training examples before starting annotation.\vspace{-5pt}

\paragraph{Annotator Recruitment and Training.}
Annotators were selected from graduate students with prior experience in machine learning or related quantitative fields. Before using the annotation platform, they completed a written tutorial on task definitions, edge cases, and examples of acceptable and unacceptable annotations; passed a 30-item calibration test spanning diverse problem types and difficulty levels with at least \(80\%\) agreement with an expert gold standard; and signed a code-of-conduct and confidentiality agreement on responsible data handling. We require all annotation experts to have at least one professional competition experience and be trained to at least a bronze medal level.\vspace{-5pt}

\paragraph{Annotation Guidelines.}
Each annotator uses an interface that displays the problem, the ground-truth final answer, the official or reference solution (when available), and one or more LLM-generated candidate rationales. When no rationale is correct, the interface instead shows three candidate incorrect rationales.
The detailed guidelines to instruct annotators are as follows:
\begin{PromptBox}{{Annotator Guidelines-P1}}
\setlength{\parindent}{0pt}
\setlength{\parskip}{0.8ex}

These guidelines define how annotators should evaluate and edit solution rationales for reasoning tasks (e.g., mathematical or logical problems). The goal is to ensure that all accepted rationales are correct, complete, and stylistically consistent.

For every candidate rationale, annotators should perform the following checks:
\begin{enumerate}[leftmargin=16pt, itemsep=0pt, topsep=0pt]
    \item \textbf{Conceptual correctness.}
    \begin{itemize}[leftmargin=16pt, itemsep=0pt, topsep=0pt]
        \item Verify that all major inference steps are logically valid.
        \item Check that all problem constraints are correctly interpreted.  
        \item Ensure that the rationale does not introduce unwarranted assumptions or ignore conditions.
    \end{itemize}
    \item[2.] \textbf{Computational correctness.}
    \begin{itemize}[leftmargin=16pt, itemsep=0pt, topsep=0pt]
        \item Check all algebraic manipulations, arithmetic calculations, and case analyses for errors.  
        \item Confirm that intermediate results are correct and consistent across steps.
    \end{itemize}
    \item[3.] \textbf{Completeness of reasoning.}
    \begin{itemize}[leftmargin=16pt, itemsep=0pt, topsep=0pt]
        \item Ensure that the solution does not skip non-trivial steps.  
        \item Include necessary intermediate results (e.g., key substitutions, simplifications, or case splits).  
        \item Check that each step is connected to the previous one and that the overall argument is coherent.
    \end{itemize}
    \item[4.] \textbf{Consistency with the final answer.}
    \begin{itemize}[leftmargin=16pt, itemsep=0pt, topsep=0pt]
        \item Confirm that the reasoning actually leads to the stated final answer.  
        \item Ensure that there is no mismatch between intermediate conclusions and the final result.  
        \item Verify that the final answer is stated explicitly, unambiguously, and in the required format.
    \end{itemize}
    \item[5.] \textbf{Clarity and style.}
    \begin{itemize}[leftmargin=16pt, itemsep=0pt, topsep=0pt]
        \item Assess whether the rationale is easy to follow for a competent reader.  
        \item Avoid unnecessary repetition, irrelevant digressions, and overly verbose explanations.  
        \item Check adherence to the agreed notation, terminology, and formatting conventions.
    \end{itemize}
\end{enumerate}
After applying the evaluation criteria above, annotators should choose exactly one of the following actions for each candidate rationale:
\begin{enumerate}[leftmargin=16pt, itemsep=0pt, topsep=0pt]
    \item \textbf{Accept with minor edits.} Use this option when the rationale is fundamentally correct and complete, but has small issues such as:
    \begin{itemize}[leftmargin=16pt, itemsep=0pt, topsep=0pt]
        \item Minor wording problems (e.g., awkward phrasing, ambiguous pronouns).
        \item Slightly unclear transitions between steps.  
        \item Cosmetic inconsistencies in notation, symbols, or formatting.  
    \end{itemize}
    In this case, annotators should directly edit the text to correct these minor issues without changing the core reasoning.

    \item[2.] \textbf{Substantive revision.}  
    Use this option when the rationale contains the correct core ideas but has more serious local problems, such as:
    \begin{itemize}[leftmargin=16pt, itemsep=0pt, topsep=0pt]
        \item Missing but recoverable intermediate steps.  
        \item Redundant detours, digressions, or unnecessary case splits.  
        \item Local mistakes (e.g., a computation error in one step, a minor mislabeling) that can be fixed without changing the overall solution strategy.  
    \end{itemize}
    \end{enumerate}
\end{PromptBox}

\begin{PromptBox}{{Annotator Guidelines-P2}}
\setlength{\parindent}{0pt}
\setlength{\parskip}{0.8ex}

\begin{enumerate}[leftmargin=16pt, itemsep=0pt, topsep=0pt]
    \item[] 
    In this case, annotators should:
    \begin{itemize}[leftmargin=16pt, itemsep=0pt, topsep=0pt]
        \item Fix computational and logical errors.  
        \item Fill in missing, non-trivial steps.  
        \item Reorganize the structure for better flow, and remove superfluous parts.  
    \end{itemize}
    \item[3.] \textbf{Complete rewrite.}
    Use this option when the rationale is not salvageable as a whole, for example:
    \begin{itemize}[leftmargin=16pt, itemsep=0pt, topsep=0pt]
        \item The overall reasoning is conceptually flawed or contradicts the problem statement.  
        \item The logical structure is inconsistent or self-contradictory.  
        \item The explanation is so unclear, disorganized, or confusing that repairing it would be harder than rewriting.  
    \end{itemize}
        In this case, annotators should:
        \begin{itemize}[leftmargin=16pt, itemsep=0pt, topsep=0pt]
            \item Discard the rationale as the primary solution.  
            \item Write a new rationale from scratch that is correct, complete, and clear.  
            \item Optionally reuse any locally correct insights from the original rationale (e.g., a correct formula, an accurate sub-case analysis), but only if they fit naturally into the new solution.
        \end{itemize}
\end{enumerate}

Annotators should choose the \textbf{least intrusive} action that yields a high-quality rationale. If in doubt between “Substantive revision” and “Complete rewrite,” prefer “Complete rewrite” when the existing structure significantly impedes clarity or correctness.
\\

Annotators should apply these guidelines consistently across all examples to ensure uniform quality and style in the final dataset.

\end{PromptBox}

\vspace{-2mm}\subsection{Details of Quality Control}\vspace{-1mm}
\label{app:quality_control}

This section provides a detailed description of the quality control procedures summarized in the main paper, including the dual-review protocol (5\$ per sample), weekly random sampling and regression testing, metric definitions, and the closed-loop feedback process for guideline refinement and model retraining.

\vspace{-2mm}\subsubsection{Dual-Review Annotation Workflow}\vspace{-1mm}

To ensure dataset quality, each problem instance (comprising the statement, associated images, and solution) underwent a dual-review protocol consisting of primary annotation followed by an independent audit. (1) A primary annotator first verified the content and assigned task-specific metadata; (2) subsequently, a blinded auditor assessed problem well-posedness, text–image alignment, and solution validity. The auditing interface presented problems in their final form, allowing reviewers to rate quality on a 5-point Likert scale and flag specific defects, such as ambiguous statements, misleading visual content, or erroneous solutions.

\vspace{-2mm}\subsubsection{Disagreement Resolution and Escalation}\vspace{-1mm}
Disagreements on critical fields, such as solution correctness or label assignment, triggered an automatic escalation to a senior reviewer. These discrepancies were identified via logical inconsistencies (e.g., opposing validity flags) or rating divergences of at least 2 points. A senior expert with domain experience then examined the full annotation history to issue a binding decision: retaining the primary annotation, adopting the auditor's revision, or rewriting the content entirely with a supporting rationale.

\vspace{-2mm}\subsubsection{Weekly Random Sampling Procedure}\vspace{-1mm}

To complement per-example dual review, we implemented weekly random sampling for quality assurance. Each week, 5\% of annotated or modified examples were randomly selected using stratified sampling across problem type, difficulty level, and source (newly created vs. revised). Senior reviewers blind to original annotator identities re-evaluated these samples.

\noindent Finally, these strategies enabled the estimation of residual error rates and enforced strict quality control. The kappa value of our annotation correctness is close to 0.86, indicating good annotation quality. Table~\ref{fig:dataset} summarizes the final coverage statistics and rationale distributions.

\vspace{-2mm}\subsection{Details of Classification Labeling}\vspace{-1mm}
\label{app:classification}

\paragraph{Taxonomy construction.}
To ensure consistent topic annotations across domains, we build a unified three-level taxonomy for OMIBench: \emph{domain} (biology, chemistry, mathematics, physics), \emph{subfield} (e.g., algebra, combinatorics, organic chemistry, mechanics), and \emph{fine-grained topic} (e.g., polynomial inequalities, graph coloring, nucleophilic substitution).\vspace{-5pt}

\paragraph{Fine-Grained Topic Annotation.}
Because the fine-grained topics are unknown a priori, we first prompt GPT-4o to perform open-ended topic analysis for each sample and to generate classification labels in the form \{CLASS: X\}. Specifically, the prompts are as follows:
\begin{PromptBox}{{Fine-Grained Topic Annotation Prompt}}
    You are an expert annotator for multi-image, multi-discipline Olympiad-level problems. Your task is to assign the most specific sub-category label to each problem.

Each sample may contain:
\begin{itemize}[leftmargin=16pt, itemsep=0pt, topsep=0pt]
    \item One or more images (diagrams, experiment setups, screenshots, etc.).
    \item Optional text (problem statement, description, notes).
\end{itemize}

Annotation rules:
\begin{itemize}[leftmargin=16pt, itemsep=0pt, topsep=0pt]
    \item Choose the label by the core method and main concept, not by surface story.
    \item Always pick the most fine-grained label available (do not annotate coarse-grained subjects like "Optics").
    \item If it mixes multiple topics, use "\#" as a divider.
    \item Use all images and text jointly; visual cues (rays, lenses, optical axes, diagrams) are as important as text.
\end{itemize}

Your output must be exactly one label string by \{CLASS: X\}. Do not output explanations or reasoning.
\end{PromptBox}\vspace{-10pt}

\paragraph{Subfield Annotation.}
We first use GPT-4o~\citep{hurst2024gpt} to assign preliminary fine-grained topic labels, and then aggregate these labels into subfield-level categories, following our taxonomy, via K-means clustering on RoBERTa~\citep{liu2019roberta} embeddings.
For each subject area, we then construct a list of valid subfield labels from the clustering results and prompt GPT-4o to select the most appropriate subfield for each previously predicted fine-grained topic.
For example, in mathematics, if the fine-grained topic is ``Integral Calculation Function Area,'' the corresponding subfield is ``Area, Perimeter, Ratios.''
The prompt template is as follows:

\begin{PromptBox}{{Subfield Annotation Prompt-P1}}
    You are an expert annotator for multi-image, multi-discipline Olympiad-level problems.

\end{PromptBox}
\begin{PromptBox}{{Subfield Annotation Prompt-P2}}
Your task is to assign the most specific sub-category label to each problem.

\texttt{[Domain name]}

Classification Labels:
\begin{itemize}[leftmargin=16pt, itemsep=0pt, topsep=0pt]
    \item \texttt{[Fine-grained topic list]}
\end{itemize}

Your output must be exactly one label string by \{CLASS: X\} from a fixed label list. Do not output explanations or reasoning.
\end{PromptBox}\vspace{-5pt}

\paragraph{Manual Verification and Correction.}
After automatic pre-labeling, all problems undergo manual review by expert annotators with relevant domain expertise. 
Annotators may retain the GPT-4o label, modify the topic within the same domain, or revise both domain and topic. 
Annotators are encouraged to propose taxonomy changes when they observe repeated, systematic mismatches between available topics and the actual problem content.

\vspace{-2mm}\section{Detailed Main Experiment}\vspace{-1mm}
\label{append:main-experiment}
\subsection{Model Inference \& Evaluation Setting}
In our main experiments, we evaluate a set of large vision-language models (LVLMs) on OMIBench in the zero-shot setting. The evaluated models include InternVL3~\citep{zhu2025internvl3}, Qwen2.5-VL~\citep{bai2025qwen25vl}, InternVL3.5~\citep{wang2025internvl35}, Qwen3-VL~\citep{bai2025qwen3vl}, GPT-4o~\citep{hurst2024gpt}, Gemini-2.5~\citep{comanici2025gemini}, OpenAI-o4-mini~\citep{openai2025o4}, GPT-5~\citep{openai2025gpt5}, and Gemini-3~\citep{gemini2025gemini3}. 
To ensure fair comparison, we standardize input prompts across models, adapting only the minimal syntax or special tokens required by each interface. The prompt specifies the task description, any associated images, and the required output format. An example prompt template is shown below:

\begin{PromptBox}{Chain-of-Thought Prompting-P1}
Please reason step by step, and then provide the final answer in the exact format: ``$\backslash$boxed\{ANSWER\}''.
\\

[Question]

{problem\_text0} [IMAGE0] {problem\_text1} [IMAGE1] ... {problem\_textn}
\\

[Choices] 
{\textcolor{gray}{\# only for multiple choices problems}}

A. {option\_A}

B. {option\_B}
...

Let's think step-by-step!
\end{PromptBox}

\vspace{-2mm}\subsection{Rationale Quality Evaluation}\vspace{-1mm}
\paragraph{Model Analyses}
We leverage the advanced reasoning capabilities of GPT-4o to assess the quality of generated rationales. Specifically, we employ two distinct sets of prompts to evaluate the intrinsic quality of the rationales and their alignment with human annotations, respectively.
To assess the intrinsic quality of the rationales, we utilize a 5-point Likert scale, prompting the model to quantify the coherence and logical validity of the reasoning process. The specific prompts are detailed below:

\begin{PromptBox}{{Rationale Quality Evaluation Prompt-P1}}
    You are given a question, a model answer, and a rationale (the step-by-step explanation or reasoning produced by the model). Your task is to evaluate the quality of the rationale, \textbf{not} the quality of the final answer itself.

Please read the rationale carefully and rate it along the following dimensions.

\subsection*{1. Logical correctness}

Does the rationale follow a sound and coherent line of reasoning?
\begin{itemize}[leftmargin=16pt, itemsep=0pt, topsep=0pt]
    \item 5 – Fully correct: Each step is logically valid, with no contradictions or clear mistakes.  
    \item 4 – Mostly correct: Overall reasoning is sound, with only minor issues that do not affect the main conclusion.  
    \item 3 – Partially correct: Some important steps are correct, but there are noticeable gaps or errors.  
    \item 2 – Mostly incorrect: Reasoning is largely flawed or inconsistent, with only a few correct fragments.  
    \item 1 – Completely incorrect: The rationale is illogical, self-contradictory, or entirely wrong.
\end{itemize}
\subsection*{2. Faithfulness to the answer}

Is the rationale genuinely explaining how the given answer is (or would be) derived, instead of being disconnected or made-up?

\begin{itemize}[leftmargin=16pt, itemsep=0pt, topsep=0pt]
    \item 5 – Fully faithful: The rationale clearly and directly supports the given answer. No hallucinated steps are needed to justify the answer.  
    \item 4 – Mostly faithful: Slight mismatches, but the core reasoning is aligned with the answer.  
    \item 3 – Partially faithful: Some parts support the answer, but other parts are irrelevant or inconsistent. 
    \item 2 – Weakly faithful: The rationale only loosely relates to the answer, or relies heavily on speculation.  
    \item 1 – Not faithful: The rationale does not explain the answer at all, or contradicts it.
\end{itemize}

\subsection*{3. Use of information}

Does the rationale appropriately use the information provided in the input (question, context, passage, etc.)?

\begin{itemize}[leftmargin=16pt, itemsep=0pt, topsep=0pt]
    \item 5 – Excellent: Uses all relevant information, does not ignore key facts, and does not add unsupported facts.  
    \item 4 – Good: Uses most important information, with minor omissions or slightly extra but harmless details.  
    \item 3 – Fair: Uses some relevant information, but misses several important points or includes noticeable unsupported content.  
    \item 2 – Poor: Rarely uses the provided information or relies heavily on invented details.  
    \item 1 – Very poor: Almost no connection to the provided information.
\end{itemize}

\subsection*{4. Clarity and readability}
Is the rationale clear, easy to follow, and understandable to a careful reader?
\begin{itemize}[leftmargin=16pt, itemsep=0pt, topsep=0pt]
    \item 5 – Very clear: Well-structured, concise, and easy to follow step by step.
\end{itemize}
\end{PromptBox}
\begin{PromptBox}{{Rationale Quality Evaluation Prompt-P2}}
\begin{itemize}[leftmargin=16pt, itemsep=0pt, topsep=0pt]
    \item 4 – Clear: Mostly easy to understand, with only minor awkward wording or small jumps.
    \item 3 – Moderately clear: Understandable overall, but contains confusing segments or disorganized structure.
    \item 2 – Unclear: Hard to follow, with long, tangled, or repetitive explanations.      
    \item 1 – Very unclear: Nearly impossible to understand the reasoning.
\end{itemize}
\subsection*{5. Level of detail}

Is the rationale appropriately detailed for explaining the answer?

\begin{itemize}[leftmargin=16pt, itemsep=0pt, topsep=0pt]
    \item 5 – Ideal detail: Includes all key steps, neither too brief nor overly long; no crucial step is skipped.  
    \item 4 – Slightly off: Missing a minor step or slightly verbose, but still adequate.  
    \item 3 – Mixed: Some important steps covered, but either too high-level or too verbose.  
    \item 2 – Inadequate: Too short and high-level, or extremely verbose without adding real value.  
    \item 1 – Very inadequate: Provides almost no meaningful explanation, or is overwhelmingly long and unfocused.
\end{itemize}
\subsection*{Overall rationale quality}

After rating each dimension, give an \textbf{overall score} for the rationale from 1 to 5, considering all aspects together:

\begin{itemize}[leftmargin=16pt, itemsep=0pt, topsep=0pt]
    \item 5 – Excellent rationale  
    \item 4 – Good rationale  
    \item 3 – Acceptable rationale  
    \item 2 – Poor rationale  
    \item 1 – Very poor rationale
\end{itemize}

Please base your judgment on the rationale text itself. Do not penalize a rationale just because the final answer is wrong, as long as the reasoning process is internally coherent and properly uses the provided information.

Your output must be exactly one label string by \{"correctness": NUMBER, "faithfulness":\vspace{5pt}

NUMBER, "information-usage": NUMBER, "clarity": NUMBER, "detail": NUMBER, "overall": NUMBER\}. Do not output explanations or reasoning.
\end{PromptBox}

Moreover, to evaluate the alignment between model-generated rationales and human-annotated ones, we prompt GPT-4o to compare the two texts and rate their similarity on a 5-point scale. The specific prompt is as follows:

\begin{PromptBox}{{Rationale Alignment Evaluation Prompt-P1}}

You are an expert evaluator specializing in natural language understanding and reasoning chains. Your task is to assess how well a model-generated rationale aligns with a human-annotated reference rationale.

Now, please evaluate the alignment between the \textbf{Model Rationale} and the \textbf{Human Reference} based on the provided \textbf{Question}. Focus on the underlying logic, the sequence of reasoning steps, and the key evidence used. Do not penalize for differences in writing style or length, provided the core logic remains identical.
\\

\textbf{Input Data:}

\textbf{[Question]}

\texttt{[Question Content]}
\\

\textbf{[Human Reference]} 

\texttt{[Human Annotated Rationale]}
\\

\textbf{[Model Rationale]} 

\texttt{[Model Predicted Rationale]}
\\

\textbf{Scoring Criteria (1-5 Likert Scale):}

\begin{itemize}[leftmargin=16pt, itemsep=0pt, topsep=0pt]
    \item \textbf{5 (Perfect Alignment):} The model rationale uses the exact same logic, key evidence, and reasoning steps as the human reference. Differences are purely stylistic.
    \item \textbf{4 (High Alignment):} The model rationale captures all key logical points of the human reference but may include minor, non-contradictory extra details or slightly different sequencing.
    \item \textbf{3 (Partial Alignment):} The model captures the main conclusion and primary evidence but misses a subordinate step or uses slightly different reasoning to arrive at the same result.
    \item \textbf{2 (Low Alignment):} The model rationale arrives at the correct answer but uses significantly different logic or misses critical evidence present in the human reference.
    \item \textbf{1 (No Alignment):} The model rationale contradicts the human reference, uses fallacious logic, or fails to address the specific constraints mentioned in the human text.\\
\end{itemize}

\textbf{Output Format:}

Provide your response in the following JSON format:

\begin{verbatim}
```json
{
  "justification": 
  "[Brief rationale]",
  "score": [Integer 1-5]
}
```
\end{verbatim}
\end{PromptBox}\vspace{-5pt}

\paragraph{Human Analyses}
To rigorously quantify the discrepancy between surface-level coherence and deep logical correctness in LVLMs, we conducted a fine-grained human evaluation. Similar to Appendix~\ref{append:manual}, we randomly sampled 100 instances from the model outputs. Unlike coarse-grained scoring, our evaluation required expert annotators to inspect the generated rationale step-by-step and identify the \textit{root cause} of the first fatal error encountered.

To decouple visual perception capabilities from reasoning engines, we developed a specific error taxonomy comprising five distinct categories: 
(1) \textbf{Visual Perception Failures}, where the model misinterprets explicit visual semantics; (2) \textbf{Cross-Image Association Failures}, indicating an inability to synthesize information across multi-view inputs; (3) \textbf{Logical Reasoning Fallacies}, covering invalid deductions and calculation errors despite correct perception; (4) \textbf{Instruction Comprehension Biases}, regarding format or constraint violations; and (5) \textbf{Other} for hallucinations or uncategorized failures. This taxonomy allows us to diagnose whether performance bottlenecks stem from the vision encoder, the cross-modal alignment, or the LLMs' reasoning core.

\begin{figure}[t]
\centering
\includegraphics[width=0.8\textwidth]{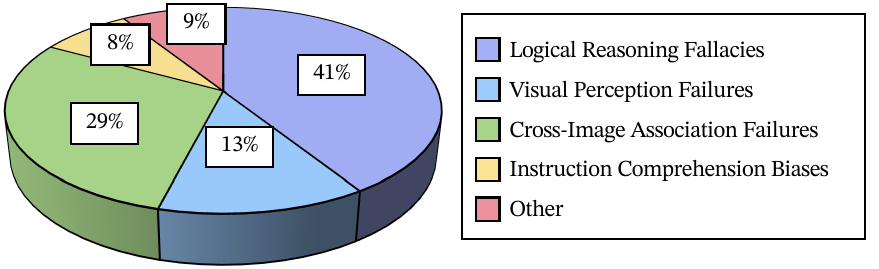}
\caption{Distribution of different reasoning error types labeled by human.}
\vspace{-12pt}
\label{fig:error}
\end{figure}

\begin{figure}
    \centering
    \vspace{-3mm}
    \includegraphics[width=0.92\textwidth]{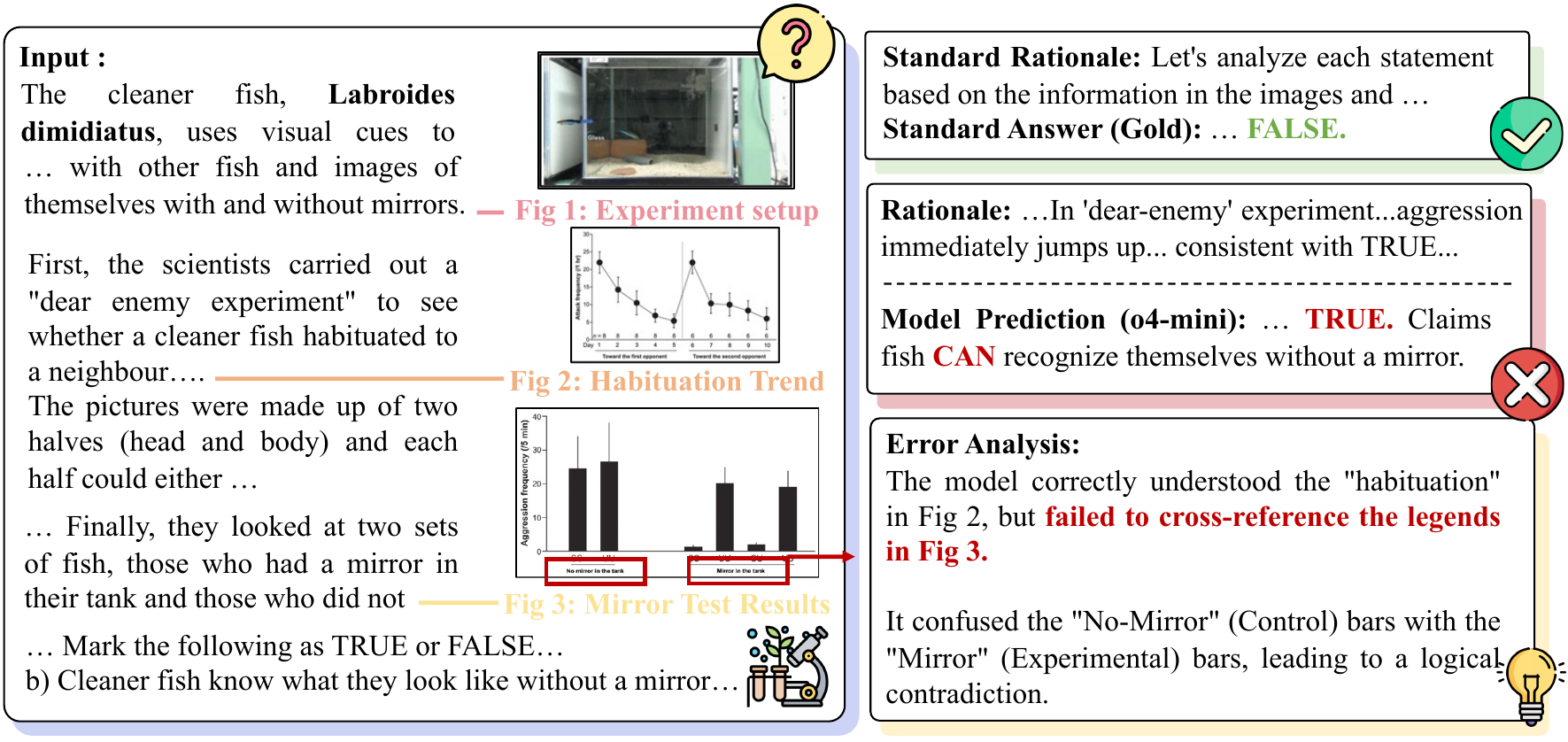} 
    \vspace{-2mm}
    \caption{Analysis of reasoning error examples on o4-mini.}
    \label{fig:case-1}
\end{figure}

\begin{PromptBox}{\textbf{Annotator Guidelines-P1}}
\setlength{\parindent}{0pt}
\setlength{\parskip}{0.8ex}
You are given a multi-image Olympiad-level problem and a model-generated rationale. Your task is to verify the correctness of the solution. If an error is found in a logical step, classify it into exactly one of the following five categories based on its primary cause.

\textbf{1. Visual Perception Failures}
The model incorrectly recognizes or misses explicit visual information within an image. This includes misreading text, misidentifying geometry/objects, or failing to perceive attributes like color or position.

\textbf{2. Cross-Image Association Failures}
The model fails to synthesize or track information across multiple images. This includes errors in understanding temporal sequences, matching objects between different views, or aggregating data from separate image panels.

\textbf{3. Logical Reasoning Fallacies}
The model correctly perceives the visual data but fails in the subsequent reasoning process. This includes invalid deductions, calculation errors, misapplication of theorems, or flawed causal logic.

\textbf{4. Instruction Comprehension Biases}
The model fails to adhere to specific constraints provided in the text prompt. This includes ignoring format requirements, violating negative constraints (e.g., "do not use calculus"), or misunderstanding the core question.

\textbf{5. Other}
Any error that does not fit the above categories. This includes hallucinating constraints that are absent from both text and images, or generating unintelligible content.
\end{PromptBox}

\begin{figure}
    \centering
    \vspace{-3mm}
    \includegraphics[width=0.92\textwidth]{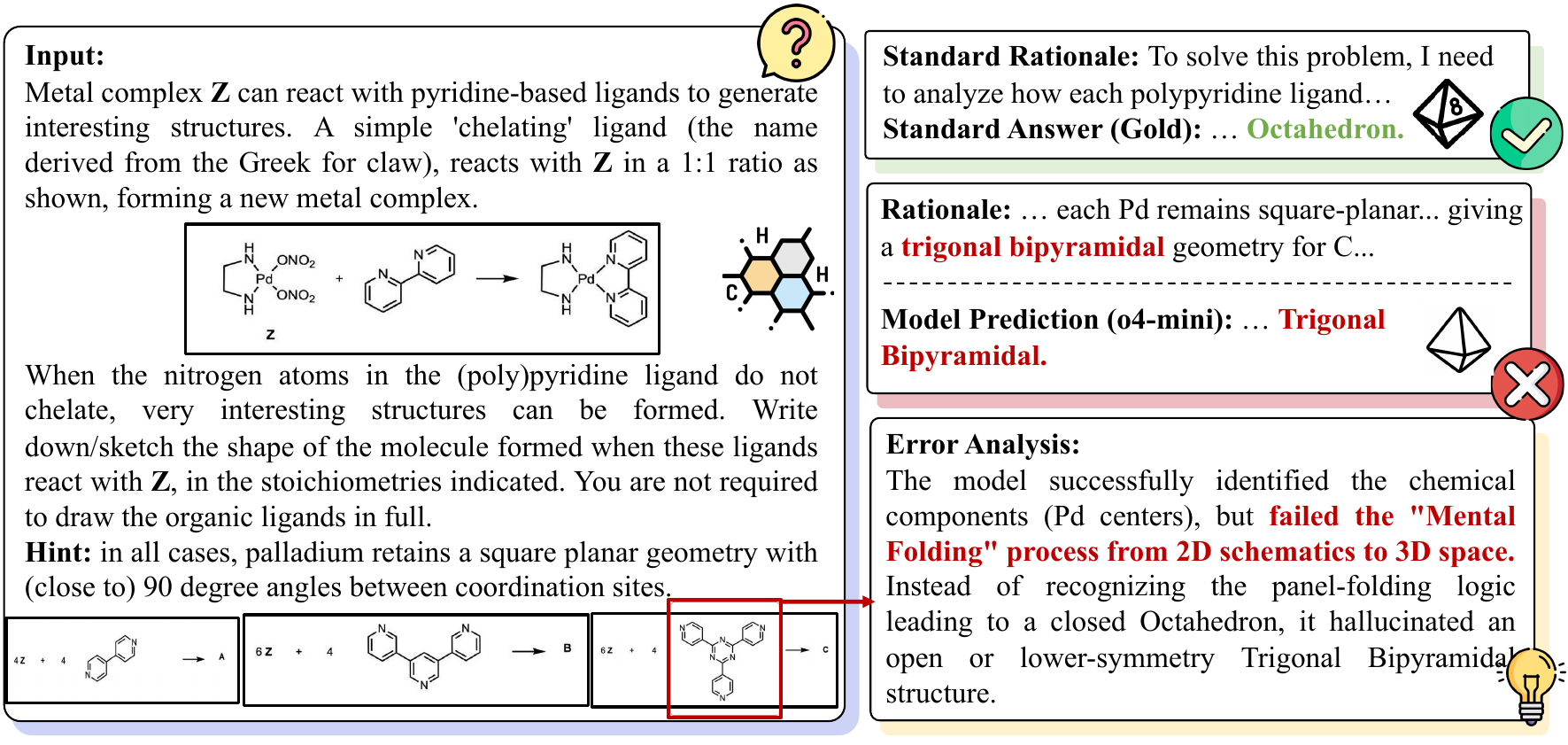} 
    \vspace{-2mm}
    \caption{Analysis of reasoning error examples on o4-mini.}
    \label{fig:case-2}
\end{figure}

\begin{figure}
    \centering
    \vspace{-3mm}
    \includegraphics[width=0.92\textwidth]{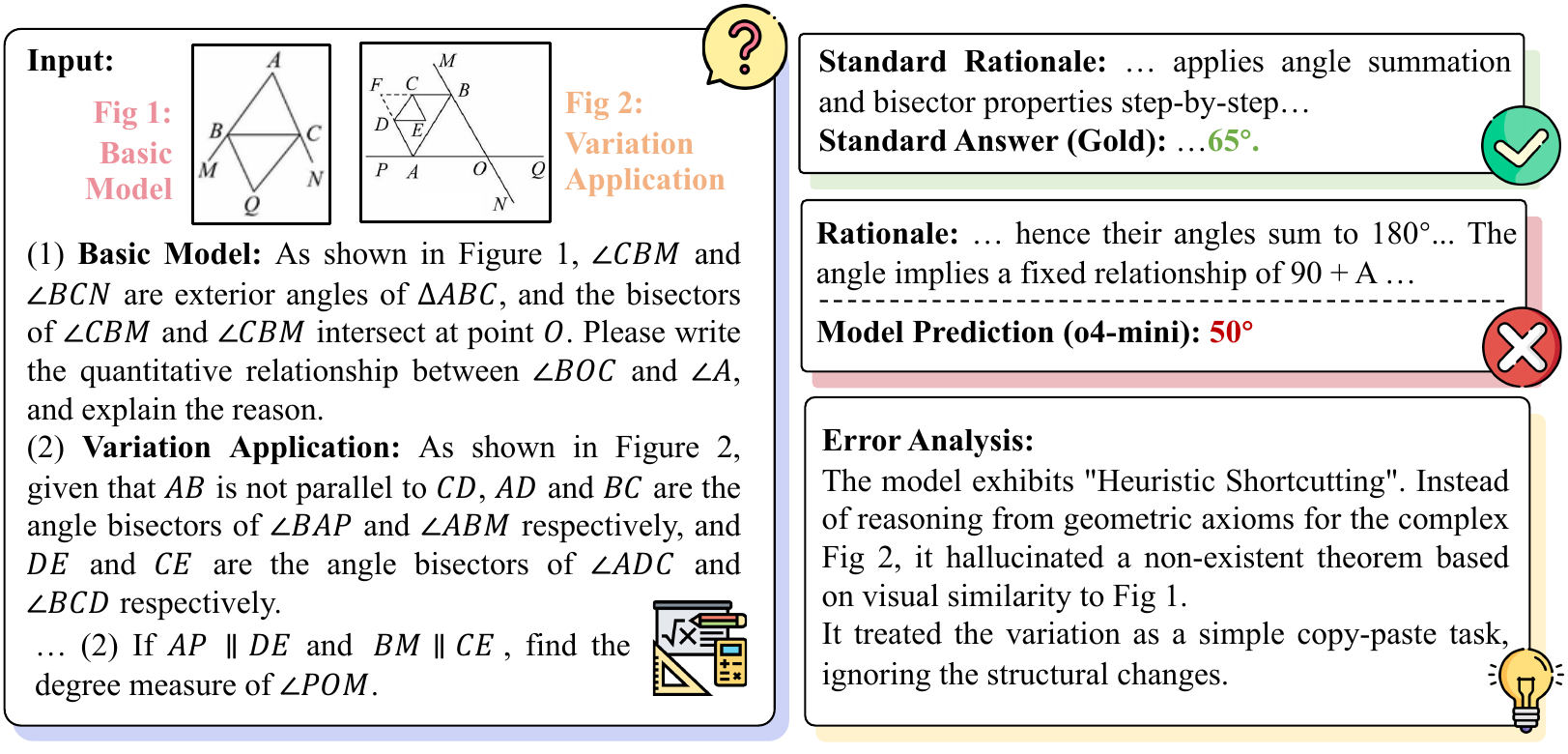} 
    \vspace{-2mm}
    \caption{Analysis of reasoning error examples on o4-mini.}
    \label{fig:case-3}
\end{figure}

\begin{figure}
    \centering
    \vspace{-3mm}
    \includegraphics[width=0.92\textwidth]{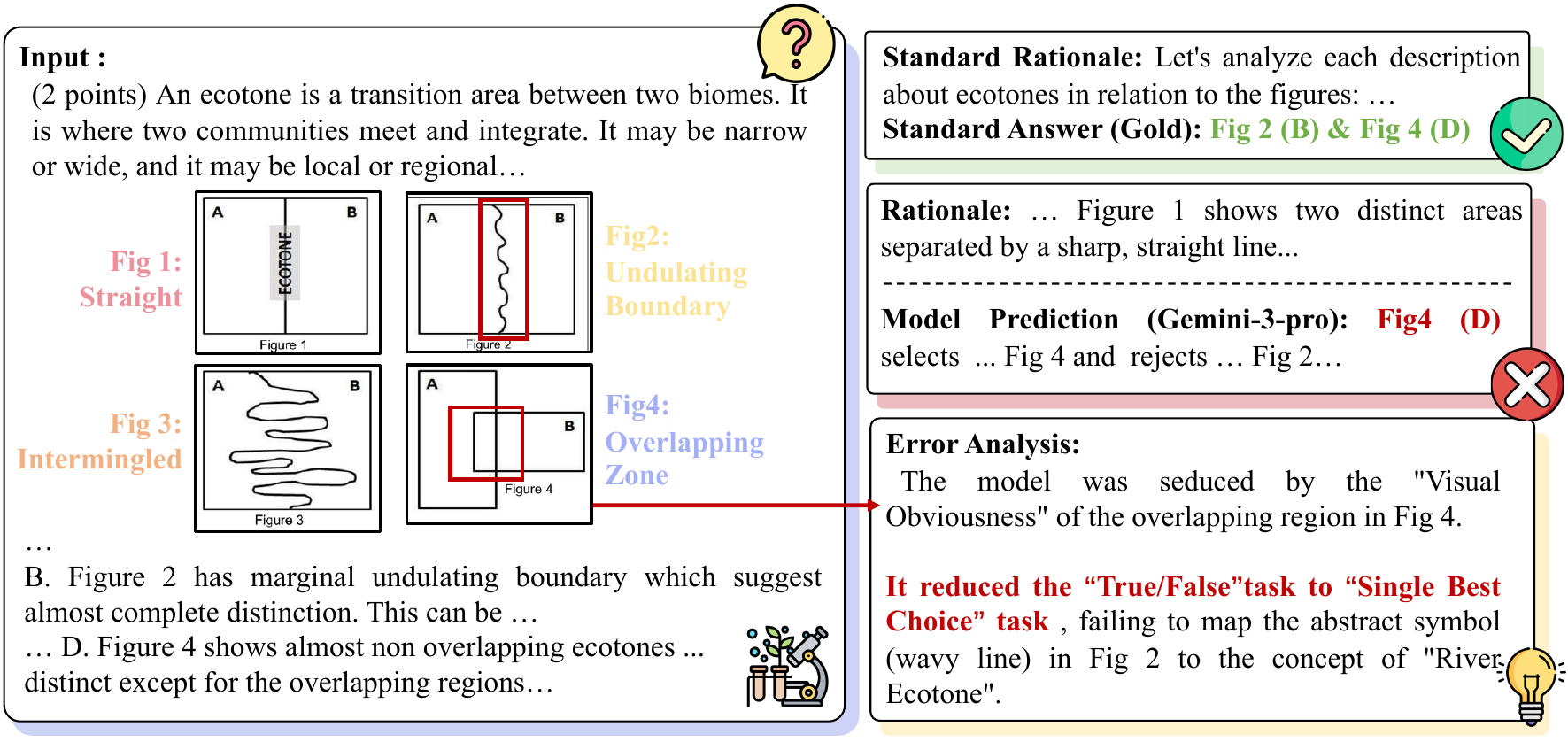} 
    \vspace{-2mm}
    \caption{Analysis of reasoning error examples on Gemini-3-Pro.}
    \label{fig:case-4}
\end{figure}

\begin{figure}
    \centering
    \vspace{-3mm}
    \includegraphics[width=0.92\textwidth]{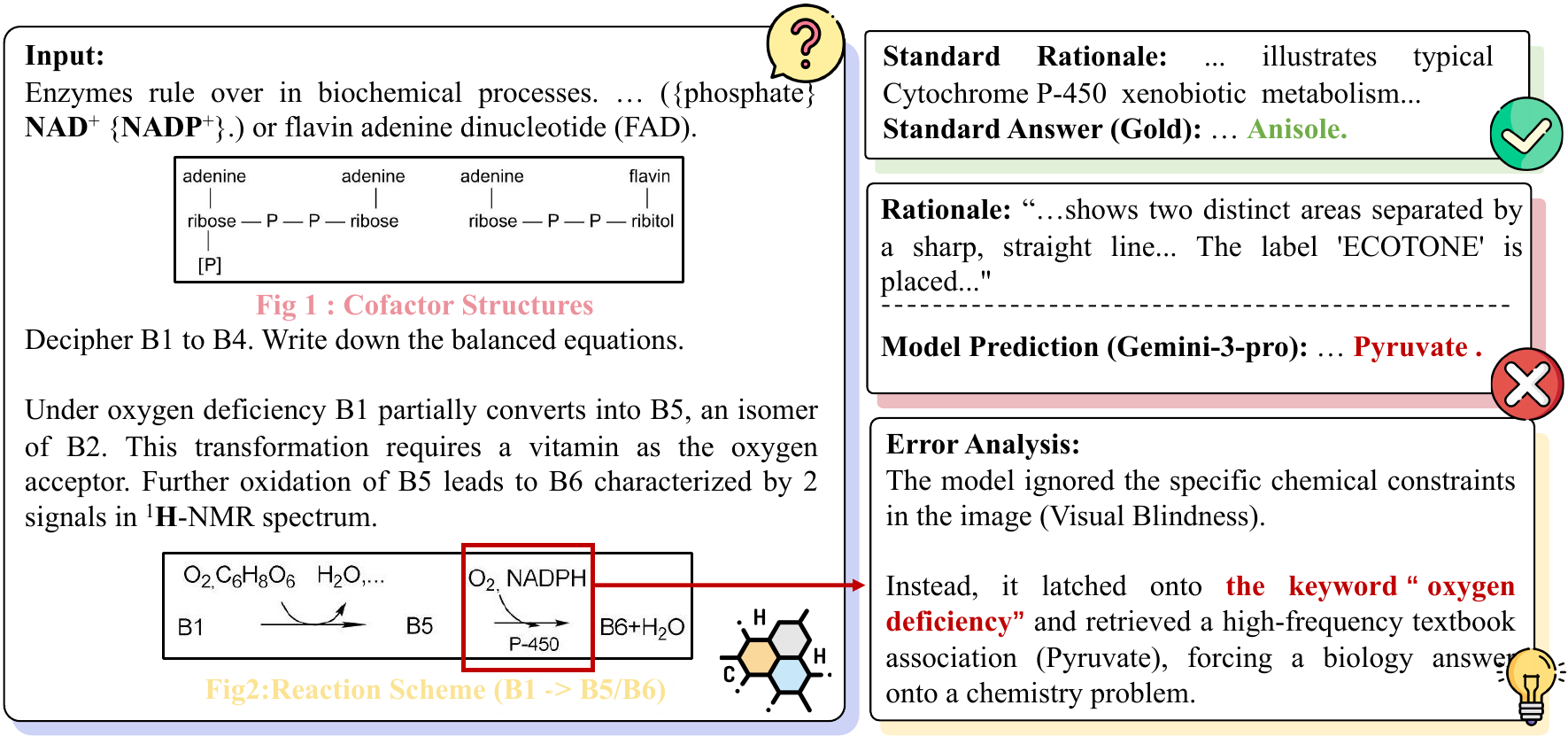} 
    \vspace{-2mm}
    \caption{Analysis of reasoning error examples on Gemini-3-Pro.}
    \label{fig:case-5}
\end{figure}

\begin{figure}
    \centering
    \vspace{-3mm}
    \includegraphics[width=0.92\textwidth]{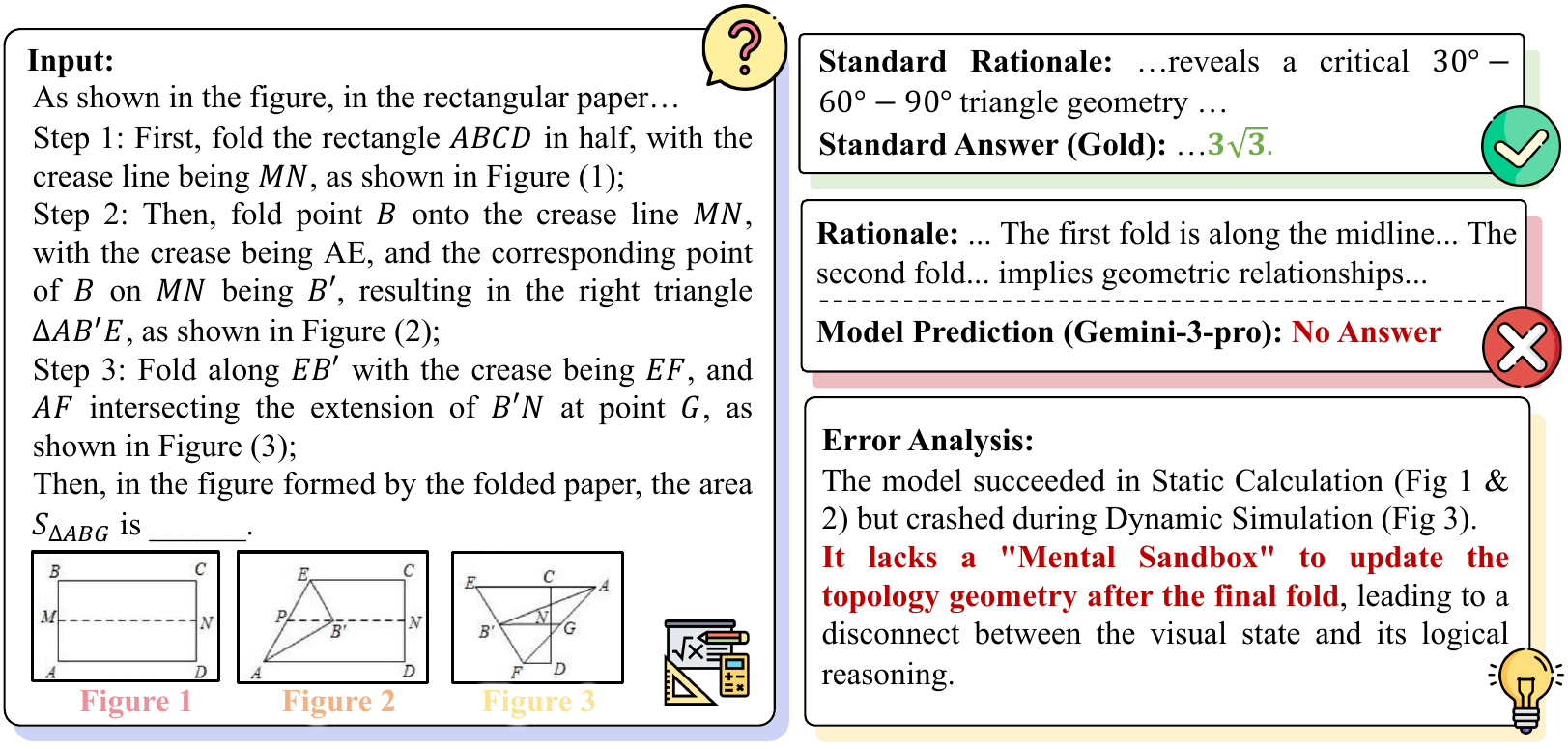} 
    \vspace{-2mm}
    \caption{Analysis of reasoning error examples on Gemini-3-Pro.}
    \label{fig:case-6}
\end{figure}

\begin{figure}
    \centering
    \vspace{-3mm}
    \includegraphics[width=0.92\textwidth]{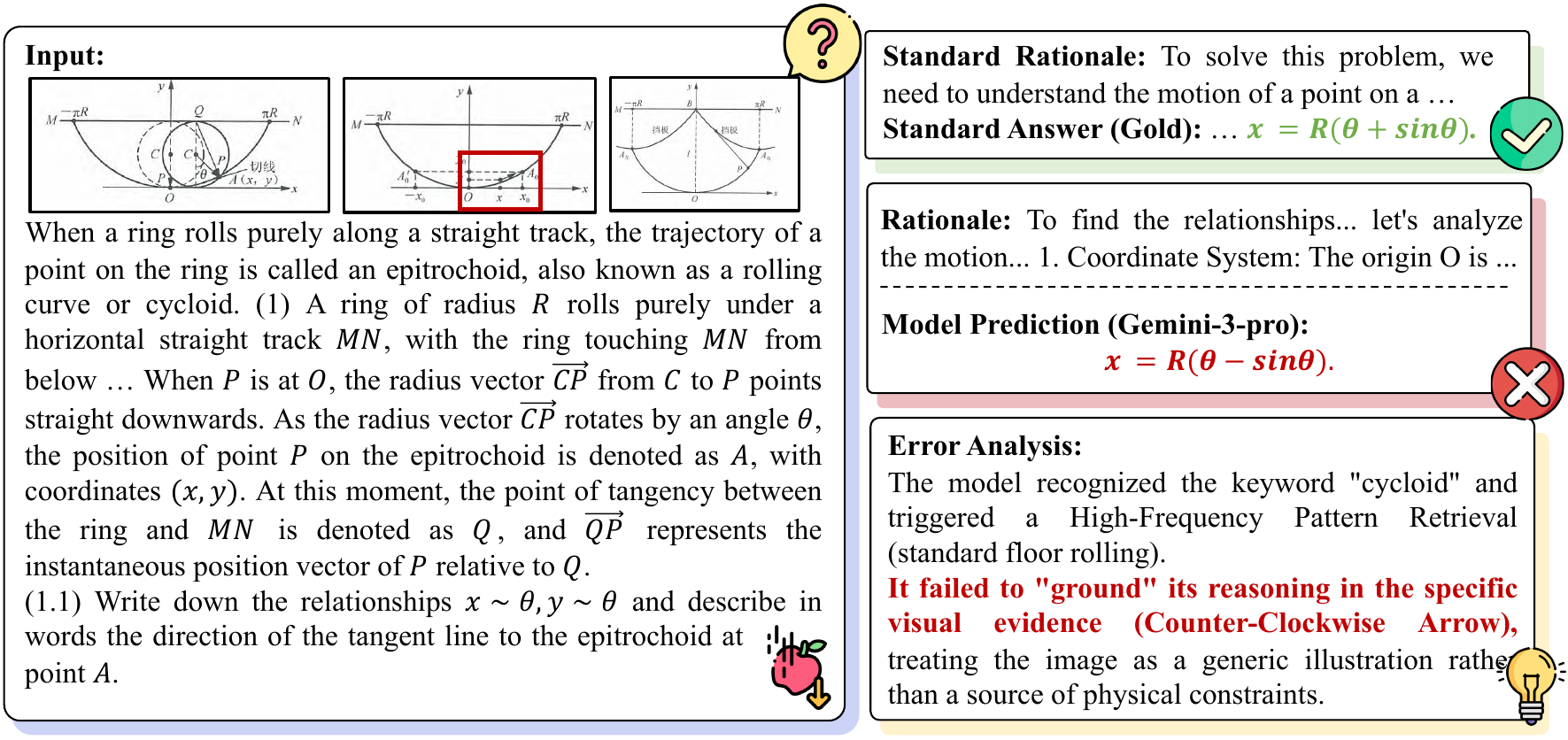} 
    \vspace{-2mm}
    \caption{Analysis of reasoning error examples on Gemini-3-Pro.}
    \label{fig:case-7}
\end{figure}

\begin{figure}
    \centering
    \vspace{-3mm}
    \includegraphics[width=0.92\textwidth]{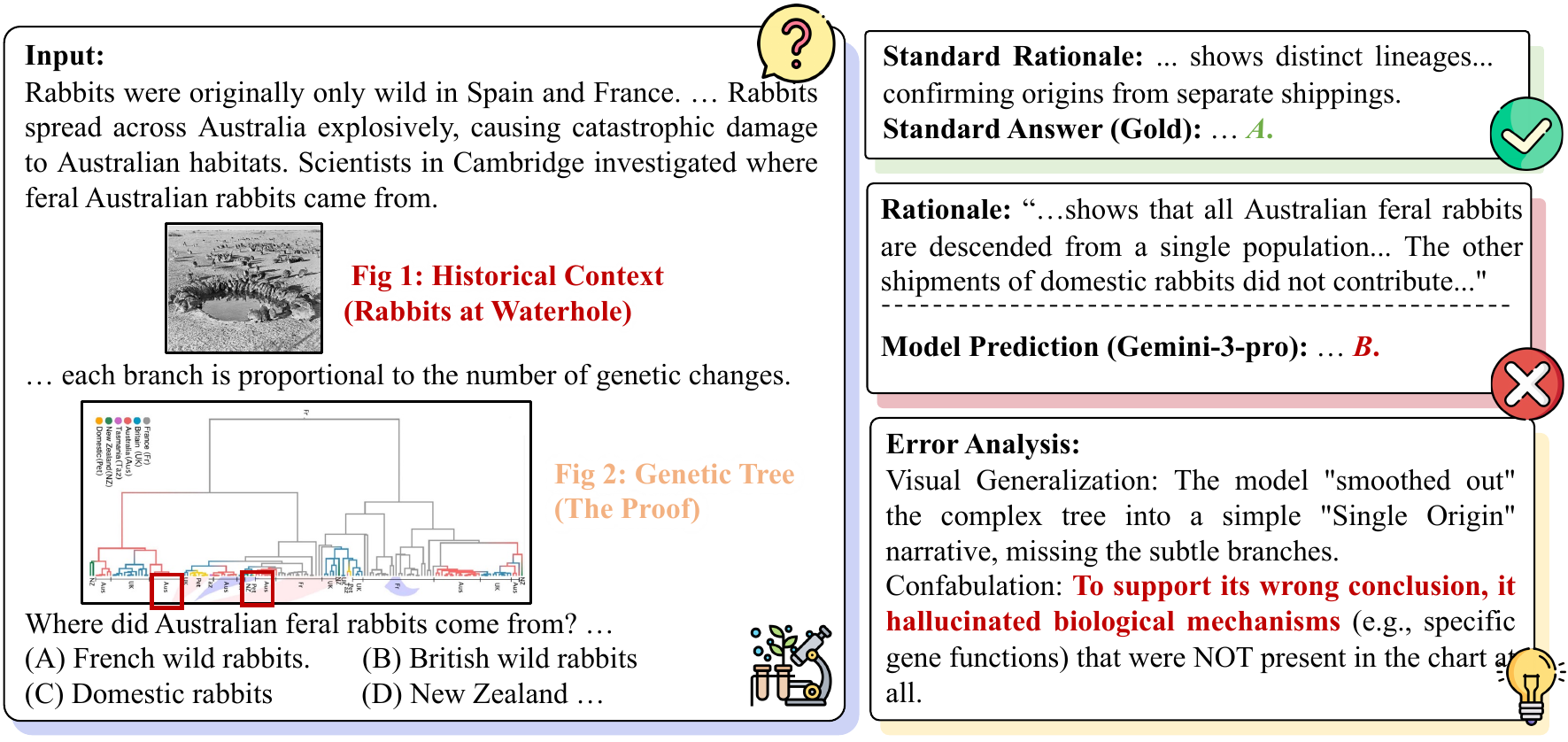} 
    \vspace{-2mm}
    \caption{Analysis of reasoning error examples on Gemini-3-Pro.}
    \label{fig:case-8}
\end{figure}

\paragraph{Case Studies of Reasoning Errors}
As illustrated in Figures~\ref{fig:case-1} to~\ref{fig:case-8}, we present representative examples of reasoning errors identified in our human evaluation. These cases highlight common pitfalls in LVLM reasoning, such as multi-image information flow confusion, misapplication of physical laws, incorrect geometric interpretations, and flawed logical deductions.



\vspace{-2mm}\section{Details of Accuracy and GPTScore}\vspace{-1mm}
\label{append:metrics}
\subsection{Matching accuracy}\vspace{-1mm}
\label{append:match-accuracy}
For each OMIBench input \(x_i\) with gold answer \(y_i\) and the model produces prediction \(\hat{y}_i\). Matching accuracy is computed at instance level:
\begin{equation}
    \text{Acc} \;=\; \frac{1}{N} \sum_{i=1}^{N} \mathbf{1}\big[\text{match}(\hat{y}_i, y_i)\big],
\end{equation}
where \(N\) denotes total evaluated instances across all tasks.

Specifically, the matching function \(\text{match}(\cdot,\cdot)\) applies consistent normalization across heterogeneous tasks:\vspace{-5pt}

\paragraph{Text normalization.} Both \(\hat{y}_i\) and \(y_i\) are lowercased, whitespace is trimmed and collapsed to single spaces, and common punctuation (periods, commas, question marks, trailing colons/semicolons) is stripped from English segments when not part of alphanumeric tokens.

Textual or equation answers. Matches are determined by longest common subsequence (LCS) ratio:
\begin{equation}
\text{match}(\hat{y}_i, y_i) = \begin{cases}
\frac{\text{Len}(\text{LCS}(\hat{y}_i, y_i))}
{\max(\text{Len}(\hat{y}_i), \text{Len}(y_i))}
& \text{if ratio} \geq \alpha \\
0 & \text{otherwise}
\end{cases},
\end{equation}
where $\text{LCS}(\cdot, \cdot)$ denotes the longest common subsequence, and $\text{Len}(\cdot)$ represents the length of given sequence. In our experiments we set \(\alpha = 75\%\).\vspace{-5pt}

\paragraph{Numeric answers.} For numeric references, we parse both \(\hat{y}_i\) and \(y_i\) into floating-point numbers after removing units and non-numeric suffixes. The match accuracy is calculated as:
\begin{equation}
\text{match}(\hat{y}_i, y_i) = 1 \quad \Leftrightarrow \quad |\hat{v}_i - v_i| \leq \epsilon,
\end{equation}
where \(v_i\) and \(\hat{v}_i\) enote parsed gold and predicted values, respectively, and \(\epsilon = 10^{-4}\). Unparseable inputs default to normalized string matching.\vspace{-5pt}

\paragraph{Multiple-choice questions.} For predefined options, we extract predictions from model outputs via regex matching of boxed notation (e.g., \(\backslash\text{boxed}\{A\}\)). A prediction is deemed correct if the extracted symbol matches either the gold label directly or, when unavailable, the corresponding option text.

\noindent Unless otherwise specified, we report micro-averaged accuracy across all evaluated tasks.

\vspace{-2mm}\subsection{Model-based GPTScore evaluation}\vspace{-1mm}

Beyond exact matching accuracy, we employ GPTScore, a semantic evaluation metric that captures partial correctness and alignment between predictions and references. This metric leverages a text-based language model (e.g., GPT-4-mini) as an automatic judge. For each input \(x_i\), gold answer \(y_i\), and model prediction \(\hat{y}_i\), we prompt the judge model using:

\begin{PromptBox}{\textbf{GPTScore Evaluation Prompt-P1}}
    You are a “Olympiad judger whose job is to decide whether two answers are equivalent in terms of their final conclusion and key reasoning.\\

Please follow these rules:
\begin{itemize}[leftmargin=16pt, itemsep=0pt, topsep=0pt]
    \item If the model’s final conclusion differs from the official answer, output “inconsistent”.
    \item Some answers have multiple answers, and missing any one is a mistake.
    \item If the final conclusion matches but the reasoning has clear logical errors or does not rigorously justify the result, output “inconsistent”.
    \item If the final conclusion matches and the reasoning is mathematically sound and sufficient to justify it, output “consistent”.
    \item Different wording, order of steps, or using a different but correct method should still be treated as “consistent”.\\
\end{itemize}

[QUESTION]

\texttt{[Multimodal Input Question]}
\\

[GOLDEN ANSWER]

\texttt{[Golden Answer]}
\\

[PREDICTED ANSWER]

\texttt{[Predicted Solution]}
\\

Output only the following format (no extra text): "ANSWER: consistent" or "ANSWER: inconsistent".
\end{PromptBox}

To compute GPTScore, we first binarize each discrete score \(s_i\) as:
\begin{equation}
\tilde{s}_i = \begin{cases}
0 & \text{if return ANSWER: consistent} \\
1 & \text{otherwise}
\end{cases},
\end{equation}

GPTScore is then the mean across all $N$ examples:
\begin{equation}
\text{GPTScore}(x_i) \;=\; \frac{1}{N} \sum_{i=1}^{N} \tilde{s}_i.
\end{equation}
For per-task results, averaging is restricted to task-specific examples. Main results report micro-averaged GPTScore across all tasks.

\vspace{-2mm}\subsection{Manual Analysis Protocol}\vspace{-1mm}
\label{append:manual}
To understand how performance diverges from GPTScore, we conducted manual analysis on cases where the two metrics disagree. For Gemini-3-Pro-Preview and InternVL3.5-1B, evaluation items were partitioned based on whether rule-based accuracy (match score>0.75 as correct) agreed with GPTScore-based labels (GPTScore=1 as correct). From disagreement cases, 100 instances per model were randomly sampled, yielding 200 instances for detailed inspection.

We collected Model Judgment and Rule Judgment labels for 200 instances and calculated agreement rates across four categories: both correct, both incorrect, Model correct but Rule incorrect, and Model incorrect but Rule correct. Two annotators with LLM evaluation experience independently assessed discordant cases, categorizing disagreement sources as: No Effective Reasoning, Reasoning Error, Rule Match Error, and Others. The annotation guideline is detailed below:

\begin{PromptBox}{{Disagreement Source Guidelines-P1}}
    
To ensure rigorous and consistent classification, we established the following definitions to characterize the primary sources of metric divergence:

\textbf{No Effective Reasoning}
\begin{itemize}[leftmargin=16pt, itemsep=0pt, topsep=0pt]
    \item \textbf{Definition:} The model outputs a final answer without providing a derivation or logical chain of thought, or the generated reasoning is incoherent/irrelevant to the query.
    \item \textbf{Key criterion:} Absence of a traceable cognitive process.
\end{itemize}
\textbf{Reasoning Error}
 \begin{itemize}[leftmargin=16pt, itemsep=0pt, topsep=0pt]
    \item \textbf{Definition:} The model attempts a step-by-step derivation but commits a logical fallacy, calculation mistake, or factual hallucination during the intermediate steps, leading to an incorrect conclusion.
    \item \textbf{Key criterion:} Flawed logic within a structured chain of thought.
\end{itemize}

\textbf{Rule Match Error}
\begin{itemize}[leftmargin=16pt, itemsep=0pt, topsep=0pt]
    \item \textbf{Definition:} The model generates a semantically correct response that aligns with the ground truth, but the rule-based evaluation (exact match) judges it as incorrect due to formatting rigidity (e.g., synonym usage, "10.0" vs. "10", or verbose phrasing).
    \item \textbf{Key criterion:} Semantic correctness rejected by syntactic rigidity.
\end{itemize}

\textbf{Others}

\begin{itemize}[leftmargin=16pt, itemsep=0pt, topsep=0pt]
    \item \textbf{Definition:} Disagreements arising from external factors such as ground-truth errors (label noise), ambiguous prompts, or unclassifiable multimodal alignment failures.
    \item \textbf{Key criterion:} Issues external to the model's reasoning capability or the extraction rule.
\end{itemize}
\end{PromptBox}

\subsection{Performance Comparison between Accuracy and GPTScore}
\begin{figure}[t]
\centering
\includegraphics[width=0.79\textwidth]{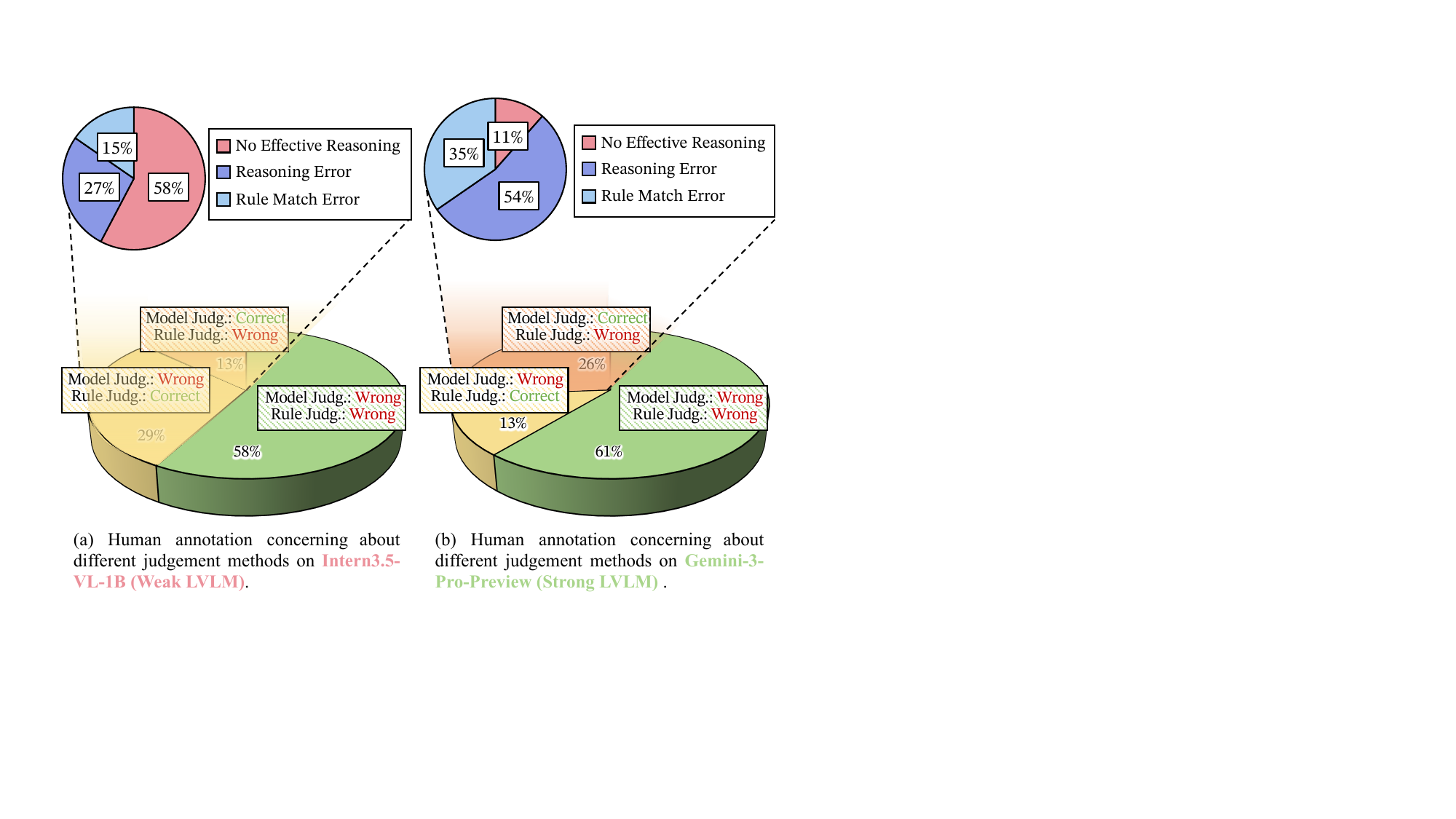}
\caption{Human annotation concerning different judgment methods (Model Judge: GPTScore, Rule Judge: Match Accuracy). Note: Correct/wrong indicates the automated method's output, not its actual accuracy.\vspace{-12pt}}
\label{fig:overestimate}
\end{figure}
\paragraph{Accuracy overestimates weaker models and underestimates stronger ones relative to GPTScore, though model ranking is preserved.}
Models with higher GPTScore also achieve higher accuracy under GPTScore-based evaluation, but accuracy systematically overestimates weaker models and underestimates stronger ones,
while largely preserving their relative ranking.
To substantiate this, we manually analyze 100 instances where accuracy and GPTScore disagree for Gemini-3-Pro-Preview and InternVL3.5-1B.

As shown in Figure~\ref{fig:overestimate}, weaker models often produce flawed reasoning yet output a definite option that rule judgment marks as correct; model judgment exposes these reasoning errors, revealing inflated accuracy. For stronger models, reasoning is typically correct, but open-ended questions lead to outputs that do not exactly match the reference; model judgment recovers these false negatives, showing that accuracy is deflated. Nonetheless, the induced model ranking remains largely consistent, indicating that accuracy is still a useful metric.

\vspace{-2mm}\section{Analyses for OMIBench Olympiad-Level Thinking Requirements}\vspace{-1mm}
\label{append:olympiad-thinking-requirement}

\paragraph{Correlation Analysis with Existing Multi-Image Benchmarks.}
To quantify how OMIBench relates to existing multi-image benchmarks while still reshaping model rankings, we perform a system-level correlation analysis with MMIU~\citep{meng2024mmiu}. To reduce reproduction cost, we reuse a subset of MMIU results from \citet{wang2025internvl35}, yielding a score pair \((s^{\text{OMI}}_m, s^{\text{MMIU}}_m)\) for each model \(m\). As reported in the main text and shown in Figure~\ref{fig:statistics}(a), this analysis yields a moderate Spearman correlation of \(\rho = 0.535\), which is well below the commonly used strong-correlation threshold of \(0.7\).

\vspace{-2mm}\section{Detailed Analyses for OMIBench Multi-Image Requirements}\vspace{-1mm}
\label{append:multi-image-requirement}
\subsection{Single- vs. Multi-image Olympiad-Level Benchmark Comparison}\vspace{-1mm}
\label{append:benchmark-comparison}
For the correlation analysis between OMIBench and OlympiadBench, we evaluate three representative LVLMs: GPT, Gemini, Qwen-VL, and InternVL.
All models are used in their official API or checkpoint form without any additional fine-tuning or task-specific adaptation.
In order to reduce evaluation cresource consumption, the partial results of OlympiadBench are taken from \citet{wan2025srpo,wang2025internvl35,team2025kimi,yue2025mimo,shen2025semi,zhang2025chainv,zha2025vision,team2025kwai,yu2025hipho}.\vspace{-5pt}

\paragraph{Spearman correlation between benchmarks.}
To study whether Olympiad-level thinking transfers from the single-image to the multi-image setting, we compute the Spearman rank correlation coefficient between model performances on OlympiadBench and OMIBench.
Concretely, for each model \(m\), we obtain its accuracy on OlympiadBench, denoted by \(a^{\text{Olympiad}}_m\), and its accuracy on OMIBench, denoted by \(a^{\text{OMI}}_m\).
We then rank the models separately according to \(a^{\text{Olympiad}}_m\) and \(a^{\text{OMI}}_m\) and apply the standard Spearman formula on these two rankings.
The resulting coefficient is \(0.614\), which is below the commonly used threshold of \(0.7\) for strong monotonic correlation,
showing that the relative ordering of LVLMs changes noticeably between the single-image and multi-image Olympiad settings.\vspace{-5pt}

\paragraph{Illustrative performance drop.}
As highlighted in the main text, a representative LVLM that reaches \(75.67\%\) accuracy on OlympiadBench drops to \(50.53\%\) accuracy on OMIBench when evaluated under the same decoding and prompting setup.
This corresponds to an absolute decrease of over \(25\%\) and a relative performance drop of approximately \(33\%\).
Together with the moderate Spearman correlation, these findings substantiate the claim that OMIBench imposes systematically stronger demands on multi-image Olympiad reasoning than its single-image counterpart.

\vspace{-2mm}\subsection{Relationship Between Number of Images and Accuracy}\vspace{-1mm}

To quantify how multi-image complexity affects model performance, we analyze accuracy as a function of the number of images per instance. Each OMIBench question is annotated with its image count \(k\). Instances are grouped into bins by \(k\) (e.g., \(k = 1, 2, 3, 4, 5, \ge 6\)), and the average accuracy for each bin is computed by aggregating predictions from all evaluated models.
Concretely, for each bin \(B_k\) with \(k\) images,  the bin-level accuracy is defined as:
\begin{equation}
\text{Acc}(k) = \frac{1}{|B_k|} \sum_{x \in B_k} \text{match}\{\hat{y}(x), y(x)\},
\end{equation}
where \(y(x)\) is the ground-truth answer for instance \(x\), \(\hat{y}(x)\) is the model prediction, and \(\text{match}\{\cdot, \cdot\}\) is the match function as mentioned in Appendix~\ref{append:match-accuracy}. Figure~\ref{fig:statistics} (b) reports the mean accuracy for each bin.

\vspace{-2mm}\subsection{Ablation: Restricting Instances to a Limited Input Image}\vspace{-1mm}

To verify that OMIBench necessitates multi-image integration rather than being solvable via limited input image cues, we constructed an ablation dataset wherein visual context was systematically restricted. For every instance containing \(K > 1\) images, we retained the original question and ground truth but supplied only the primary image (the first in the canonical sequence). This approach isolates the impact of visual context reduction while maintaining the distributional properties of the query set.

We evaluated model performance across two conditions: (1) the standard multi-image setting, where models access the full visual complement (\(k=K\) images); and (2) the partial-image ablation, where input is strictly limited to \(k<K\) images per instance. 
As illustrated in Figure~\ref{fig:statistics} (c), limiting visual input precipitates a marked decline in performance. Across all evaluated LVLMs, the mean accuracy drops by at least \(10\%\) in the single-image setting relative to the multi-image baseline. The degradation is particularly pronounced for inquiries requiring cross-referencing or comparative analysis, confirming that high performance on OMIBench relies on joint reasoning over multiple visual inputs.

\vspace{-2mm}\section{Detailed Analyses for “Combined Multi-Image and Olympiad-Level Thinking”}\vspace{-1mm}
\label{append:combined-requirement}
\subsection{Correlation with MMIU and OlympiadBench}\vspace{-1mm}

Figure~\ref{fig:statistics} (a) examines model ranking consistency. MMIU prioritizes multi-image perception, whereas OlympiadBench emphasizes single-image reasoning; OMIBench integrates these demands by requiring reasoning across multiple images. Empirically, rankings between MMIU and OlympiadBench diverge, confirming their disparate foci. Conversely, OMIBench correlates more strongly with each baseline than the baselines do with one another. This suggests OMIBench effectively bridges the two domains, capturing the joint capabilities of multi-image understanding and complex problem-solving.

\vspace{-2mm}\subsection{Detailed Error Analysis}\vspace{-1mm}
\label{app:error_analysis}

In this section, we formally define the annotation protocol, the taxonomy of failure types, and the qualitative patterns observed for each category on OMIBench. The goal is to make the reported percentages in Figure~\ref{fig:error} reproducible and interpretable.

\vspace{-2mm}\subsubsection{Annotation Protocol}\vspace{-1mm}

We analyzed OMIBench examples where LVLM responses fell below a predefined correctness threshold under GPTScore. These \emph{candidate errors} underwent a rigorous secondary inspection to elucidate failure mechanisms. For each instance, annotators examined the visual inputs, instructions, reference solutions, and model outputs. A prerequisite validation step excluded false positives where GPTScore misclassified valid responses. Confirmed errors were then assigned a \emph{primary} failure mode based on the fundamental deficit preventing a correct output.

Figure~\ref{fig:error} illustrates the resulting distribution: visual perception failures (35\%), cross-image association failures (30\%), logical reasoning fallacies (25\%), and instruction comprehension biases (10\%). The following subsections detail the operational definitions of each category.\vspace{-5pt}

\paragraph{Visual Perception Failures} occur when the LVLM misperceives basic visual facts in one or more images, yielding wrong or incomplete scene descriptions, even for simple questions, and thus unreliable downstream reasoning. 
We label an error as a visual perception failure when \emph{at least one} of the following holds:
\begin{enumerate}[leftmargin=16pt, itemsep=0pt, topsep=0pt]
\item \textbf{Object misrecognition:} The model assigns an incorrect object category that the question depends on (e.g., bus vs.\ truck, dog vs.\ cat).
\item \textbf{Attribute misclassification:} The model gets the object class right but misperceives salient attributes such as color, number, relative size, pose, or state (e.g., open vs.\ closed, full vs.\ empty).
\item \textbf{Spatial relation errors:} The model misreads coarse spatial relations or layout (e.g., left vs.\ right, in front of vs.\ behind, above vs.\ below) that are visually clear and explicitly queried.
\item \textbf{Salient detail omission:} The answer would be correct if a clearly visible but critical detail did not exist (e.g., a small but prominent symbol, icon, or text overlay).
\end{enumerate}\vspace{-5pt}

\paragraph{Cross-Image Association Failures} occur when the LVLM parses each image reasonably well in isolation but fails to correctly relate them when the question requires comparing, contrasting, or aggregating information across images.
We assign this label when:
\begin{enumerate}[leftmargin=16pt, itemsep=0pt, topsep=0pt]
\item The model’s descriptions of individual images (paraphrased or inferred from its answer) are largely accurate.
\item The question explicitly or implicitly involves multiple images (e.g., ``between the first and second image'', ``across all panels'').
\item The error stems from misalignment, confusion, or omission in how information from different images is combined.
\end{enumerate}\vspace{-5pt}

\paragraph{Logical Reasoning Fallacies} occur when the LVLM’s basic visual understanding and cross-image mapping are adequate, but the chain of reasoning leading to the final answer contains flawed logical steps; fixing the reasoning alone would yield the correct answer.
We annotate an error as a logical reasoning fallacy when:
(1) The model’s implicit or explicit description of relevant visual facts is broadly correct; perception is not the main source of error.
(2) The natural-language explanation, if present, shows misapplied inference rules, unsupported assumptions, or inconsistent intermediate conclusions.
(3) Adjusting the reasoning alone, without changing the perceived facts, would fix the answer.

Error distributions highlight complementary weaknesses in LVLM. Visual perception (35\%) and cross-image association failures (30\%) indicate persistent limits in fine-grained visual understanding and multi-image integration, while logical reasoning (25\%) and instruction comprehension errors (10\%) show that stronger visual encoders alone are insufficient without advances in structured reasoning and multimodal instruction following.

These results motivate future work on: (i) stronger low- and mid-level visual representations, (ii) explicit cross-image alignment and aggregation mechanisms, (iii) more reliable, verifiable reasoning procedures, and (iv) training schemes that sharpen sensitivity to multimodal instructions and output constraints.

\vspace{-2mm}\section{Details for Long CoT Experiments}\vspace{-1mm}
\label{app:long-cot-details}

This section provides the full experimental details for Section~\ref{sec:long-cot} (\emph{Can Long Chain-of-Thought Strategies Help?}), including the exact prompting templates, decoding configurations, and the definition of the ``thinking'' and ``no-thinking'' settings used on OMIBench.

\vspace{-2mm}\subsection{Prompting strategies on OMIBench}\vspace{-1mm}
\label{app:long-cot:prompt-strategies}

To provide the exact templates used for the comparison in Figure~\ref{fig:thinking} (a), we systematically evaluate widely-used Chain-of-Thought prompting strategies on OMIBench, including Least-to-Most~\citep{zhou2023leasttomost}, Plan-and-Solve~\citep{wang-etal-2023-plan}, VoT~\citep{we2024vot}.
All prompts follow the generic instruction format below:\vspace{-5pt}

\paragraph{Least-to-Most Prompting~\citep{zhou2023leasttomost}} is a strategy that decomposes complex problems into a sequence of simpler subproblems, solving them one by one. Specifically, the prompt used is as follows:
\begin{PromptBox}{Least-to-Most Prompting}
Please reason step by step, and then provide the final answer in the exact format: $\backslash$boxed\{ANSWER\}.
\\

[Question]

{problem\_text0} [IMAGE0] {problem\_text1} [IMAGE1] ... {problem\_textn}
\\

[Choices] 
{\textcolor{gray}{\# only for multiple choices problems}}

A. {option\_A}

...

Let’s break down this problem and solve it one by one.
\end{PromptBox}\vspace{-5pt}

\paragraph{Plan-and-Solve Prompting~\citep{wang-etal-2023-plan}} first devises a high-level plan to tackle the problem, then executes the plan step by step. The specific prompt used is as follows:
\begin{PromptBox}{Plan-and-Solve Prompting}
Please reason step by step, and then provide the final answer in the exact format: $\backslash$boxed\{ANSWER\}.
\\

[Question]

{problem\_text0} [IMAGE0] {problem\_text1} [IMAGE1] ... {problem\_textn}
\\

[Choices] 
{\textcolor{gray}{\# only for multiple choices problems}}

A. {option\_A}

...

Let’s first understand the problem and devise a plan to solve it. Then, let’s carry out the plan and solve the problem step by step.
\end{PromptBox}\vspace{-5pt}

\paragraph{Visualization-of-Thought (VoT) Prompting~\citep{we2024vot}} encourages the model to visualize intermediate states after each reasoning step to enhance clarity and understanding. The specific prompt used is as follows:
\begin{PromptBox}{Visualization-of-Thought (VoT) Prompting}
Please reason step by step, and then provide the final answer in the exact format: $\backslash$boxed\{ANSWER\}.
\\

[Question]

{problem\_text0} [IMAGE0] {problem\_text1} [IMAGE1] ... {problem\_textn}
\\

[Choices] 
{\textcolor{gray}{\# only for multiple choices problems}}

A. {option\_A}

B. {option\_B}

...

Visualize the state after each reasoning step.
\end{PromptBox}

\vspace{-2mm}\subsection{``Thinking'' vs.\ ``No-Thinking'' Prompting Modes}\vspace{-1mm}
\label{app:long-cot:thinking-modes}

We next describe how the ``thinking'' and ``no-thinking'' prompting modes in Fig.~\ref{fig:thinking}(b) are implemented for both reasoning-oriented and non-reasoning LVLMs.
In the "no-thinking" mode, the prompt instructs the model to avoid step-by-step reasoning and output only the final answer:
\begin{PromptBox}{No-Think Prompting}
Return only the final answer in this exact format: $\backslash$boxed\{ANSWER\}.
\\

[Question]

{problem\_text0} [IMAGE0] {problem\_text1} [IMAGE1] ... {problem\_textn}
\\

[Choices] 
{\textcolor{gray}{\# only for multiple choices problems}}

A. {option\_A}

B. {option\_B}
...
\end{PromptBox}

\begin{table}[t]
\centering
\scriptsize
\setlength{\tabcolsep}{2.4pt}
\resizebox{\textwidth}{!}{
    \begin{tabular}{lcccccccccc}
\toprule
\multirow{2}{*}{\textbf{Model}} & \multicolumn{2}{c}{\textbf{Biology}} & \multicolumn{2}{c}{\textbf{Chemistry}} & \multicolumn{2}{c}{\textbf{Mathematics}} & \multicolumn{2}{c}{\textbf{Physics}} & \multicolumn{2}{c}{\textbf{Total}} \\
\cmidrule{2-11}
 & ACC & Score & ACC & Score & ACC & Score & ACC & Score & ACC & Score \\
\midrule
Qwen3-VL-2B-Instruct & 27.44  & 19.92  & 12.33  & 5.07  & 11.53  & 6.99  & 12.50  & 8.75  & 14.99  & 9.69 \\
Qwen3-VL-2B-Thinking     & 24.86  & 9.56  & 10.46  & 1.38  & 22.25  & 15.81  & 10.75  & 6.84  & 17.12  & 9.38 \\
\midrule
Qwen3-VL-4B-Instruct & 43.03  & 36.65  & 17.05  & 11.06  & 27.91  & 23.72  & 18.40  & 13.92  & 25.95  & 20.95 \\
Qwen3-VL-4B-Thinking     & 43.58  & 36.65  & 18.94  & 11.52  & 50.80  & 44.65  & 20.41  & 20.75  & 34.45  & 30.03 \\
\midrule
Qwen3-VL-8B-Instruct & 46.61  & 43.43  & 16.13  & 17.05  & 27.44  & 29.30  & 20.05  & 18.63  & 26.85  & 26.55 \\
Qwen3-VL-8B-Thinking     & 52.39  & 52.59  & 15.96  & 17.05  & 49.90  & 49.30  & 21.97  & 30.66  & 35.84  & 38.65 \\
\midrule
Qwen3-VL-30B-A3B-Instruct & 48.51  & 48.61  & 13.29  & 12.90  & 31.42  & 32.33  & 20.02  & 20.99  & 28.03  & 28.59 \\
Qwen3-VL-30B-A3B-Thinking   & 54.44  & 55.78  & 28.13  & 32.26  & 63.70  & 59.30  & 27.92  & 37.50  & 44.63  & 47.20 \\
\midrule
Qwen3-VL-32B-Instruct & 57.62  & 58.57  & 14.09  & 20.74  & 44.40  & 40.70  & 25.48  & 25.00  & 35.87  & 35.78 \\
Qwen3-VL-32B-Thinking     & 65.72  & 64.14  & 29.83  & 24.88  & 60.66  & 60.93  & 31.92  & 41.98  & 47.34  & 49.54 \\
\midrule
Qwen3-VL-235B-A22B-Instruct & 60.41  & 63.20  & 17.23  & 22.58  & 37.48  & 34.19  & 23.77  & 23.58  & 34.11  & 34.39 \\
Qwen3-VL-235B-A22B-Thinking  & 55.00  & 61.35  & 34.14  & 33.18  & 53.71  & 48.84  & 26.81  & 43.87  & 42.12  & 47.05 \\
\bottomrule
\end{tabular}
}
\caption{``Thinking'' and ``Instruct'' results on OMIBench, where the bold content denotes the best performance in each category.}
\label{tab:thinking-mode}
\end{table}

\vspace{-2mm}\subsection{``Thinking'' vs.\ ``Instruct'' Model Variants}\vspace{-1mm}
\label{app:long-cot:qwen3-details}

Finally, we describe the usage of the Qwen3-VL ``thinking'' and ``instruct'' variants whose comparison is summarized in Figure~\ref{fig:thinking}(c).
We use two official checkpoints: ``Qwen3-VL-Instruct,'' an instruction-following vision-language model optimized for general-purpose multimodal tasks; and ``Qwen3-VL-Thought,'' a variant optimized for long-form reasoning that supports an explicit ``thinking'' mode with extended internal deliberation.
As shown in Figure~\ref{fig:thinking} (c), although the gains remain below Olympiad-level performance, they constitute the largest relative improvement among all tested Long CoT paradigms, indicating that specialized long-reasoning training can partially enhance multimodal Olympiad performance.

\vspace{-2mm}\section{Detailed Protocols for Test-Time Scaling Experiments}\vspace{-1mm}
\label{append:tts-exp}
Unless otherwise specified, all results are reported on the OMIBench test split using GPTScore, and each configuration is evaluated on the full benchmark.

\vspace{-2mm}\subsection{Sequential Scaling Protocol}\vspace{-1mm}

To study sequential test-time scaling, we vary the maximum number of newly generated tokens per example, \(L_{\max}\), while keeping all other hyperparameters fixed and re-evaluating the full OMIBench test set for each configuration. We sweep
\begin{equation}
    L_{\max} \in \{512, 1{,}024, 2{,}048, 4{,}096, 8{,}192, 16{,}384\},
\end{equation}
a near-geometric progression that provides dense coverage on a log scale with a manageable number of runs.

\vspace{-2mm}\subsection{Parameter Scaling Protocol}\vspace{-1mm}

To analyze how test-time scaling interacts with model size, we evaluate two families of open-source multimodal language models: InternVL and QwenVL. Within each family, we use checkpoints spanning roughly \(1\)B to \(235\)B parameters (see Table~\ref{tab:main}).
The resulting performance–parameter-count curves are shown in Figure~\ref{fig:scaling} (b), with parameter counts on a logarithmic scale. We mark the saturation region as the smallest parameter size beyond which all larger models yield less than 0.5 absolute GPTScore improvement.
InternVL improves up to approximately the mid-sized checkpoints and then saturates around \(25\%\) GPTScore, whereas QwenVL continues to improve up to its largest public variant, plateauing around \(35\%\) GPTScore.

\vspace{-2mm}\subsection{Parallel Scaling Protocol}\vspace{-1mm}

To evaluate parallel test-time scaling, we fix the model and prompting scheme and vary the number of independent samples per example, denoted by \(k\). For each configuration, we draw \(k\) stochastic reasoning trajectories and aggregate them via majority vote over the final answers. We sweep
\begin{equation}
k \in \{1, 3, 4, 8, 16\}.
\end{equation}
For \(k = 1\), this reduces to standard single-sample decoding. For \(k > 1\), we keep all per-sample decoding hyperparameters fixed and only change the number of parallel draws. Since self-consistency is not well-defined for \(k = 2\), we use \(k = 3\) as the smallest multi-sample setting.

To enable self-consistency, we decode with temperature \(T = 0.6\) and a maximum reasoning length \(L_{\max} = 16{,}384\) tokens per sample. After decoding, we apply the common post-processing pipeline to extract and normalize the final answer from each of the \(k\) samples and then take the majority-voted answer.
For each \(k\), we compute the mean GPTScore on the full OMIBench test set using the majority-voted predictions, yielding the accuracy–sampling curve in Figure~\ref{fig:scaling}(c). Plotting GPTScore against \(\log_2(k)\) reveals an approximately log-linear relationship between the number of samples and performance over the examined range of \(k\).

\vspace{-2mm}\subsection{Details of In-Context Learning Experiments on OMIBench}\vspace{-1mm}
\label{app:icl-details}

This section details the experimental protocol for the in-context learning (ICL) results reported in Figure~\ref{fig:think-with-image}(a), including the construction of in-context examples, the definition of the No-Image-ICL, Single-Image-ICL and Multi-Image-ICL conditions, and the control choices used to ensure fair comparison across conditions.\vspace{-5pt}

\paragraph{Prompt template and formatting.}
For each test instance, we randomly sample $k$ demonstrations from OMIBench; these source problems are excluded when computing the evaluation metrics. For each instance, the sampled demonstrations are fixed across all ICL conditions to ensure a fair comparison.
All ICL variants share a unified prompt template to isolate the effect of visual context.
Each prompt is structured as follows:
\begin{PromptBox}{{In-Context Learning Prompt Template-P1}}

Please reason step by step, and then provide the final answer in the exact format: ``$\backslash$boxed\{ANSWER\}''.
\\

[EXAMPLE 1]

[Question]

\texttt{[Problem Text 0] [IMAGE0] ... [Problem Text $n_K$]}

[Choices]
{\textcolor{gray}{\# only for multiple choices problems}}

[Solution]

\texttt{[Example Solution]}
\\

[EXAMPLE 2]

[Question]

\texttt{[Problem Text 0] [IMAGE0] ... [Problem Text $n_K$]}

[Choices] 
{\textcolor{gray}{\# only for multiple choices problems}}

[Solution]

\texttt{[Example Solution]}

...\\

[REQUEST]

[Question]

\texttt{[Problem Text 0] [IMAGE0] ... [Problem Text $n_K$]}

[Choices] 
{\textcolor{gray}{\# only for multiple choices problems}}

A. {option\_A}

B. {option\_B}

...
\end{PromptBox}\vspace{-5pt}

\paragraph{No-Image-ICL Configuration.}
The No-Image-ICL condition tests whether models can exploit purely textual patterns in the demonstrations, even though OMIBench is a multimodal benchmark.
Concretely, for each of the \(k\) demonstrations, we remove all images from the model input but keep the textual problem description (including any references to images, image indices, or regions) and the answer line.
Thus, the model can only rely on textual information in the demonstrations, but still has access to the full visual information for the target question. 
As reported in the main text, adding these purely textual in-context examples substantially improves OMIBench performance compared to the zero-shot baseline.\vspace{-5pt}

\paragraph{Single-Image-ICL Configuration.}
The Single-Image-ICL condition evaluates whether attaching a \emph{single} representative image to each demonstration offers additional benefits over text-only demonstrations.
For each of the \(k\) demonstration instances:
If the original OMIBench question contains a single image, we attach that image to the demonstration, exactly as in the dataset.
All other aspects of the prompt (textual problem statement, options, and answer format) are identical to the No-Image-ICL case, and the test instance is again provided with its full set of images.
This condition probes whether current LVLMs can effectively exploit minimal visual context in the demonstrations to further improve over text-only ICL.\vspace{-5pt}

\paragraph{Multi-Image-ICL Configuration.}
The Multi-Image-ICL condition provides the model with the full visual complexity of OMIBench within the in-context examples.
For each of the \(k\) demonstrations, we attach \emph{all} images associated with that OMIBench instance, preserving the original ordering.
Thus, demonstrations in this condition mirror the multimodal structure of the test instance itself.
The textual content and answer format again remain unchanged relative to the other ICL configurations, so that any performance differences can be attributed to how well the model leverages multi-image visual context in the demonstrations.

\begin{figure}[t]
    \centering
    \vspace{-3mm}
    \includegraphics[width=0.92\textwidth]{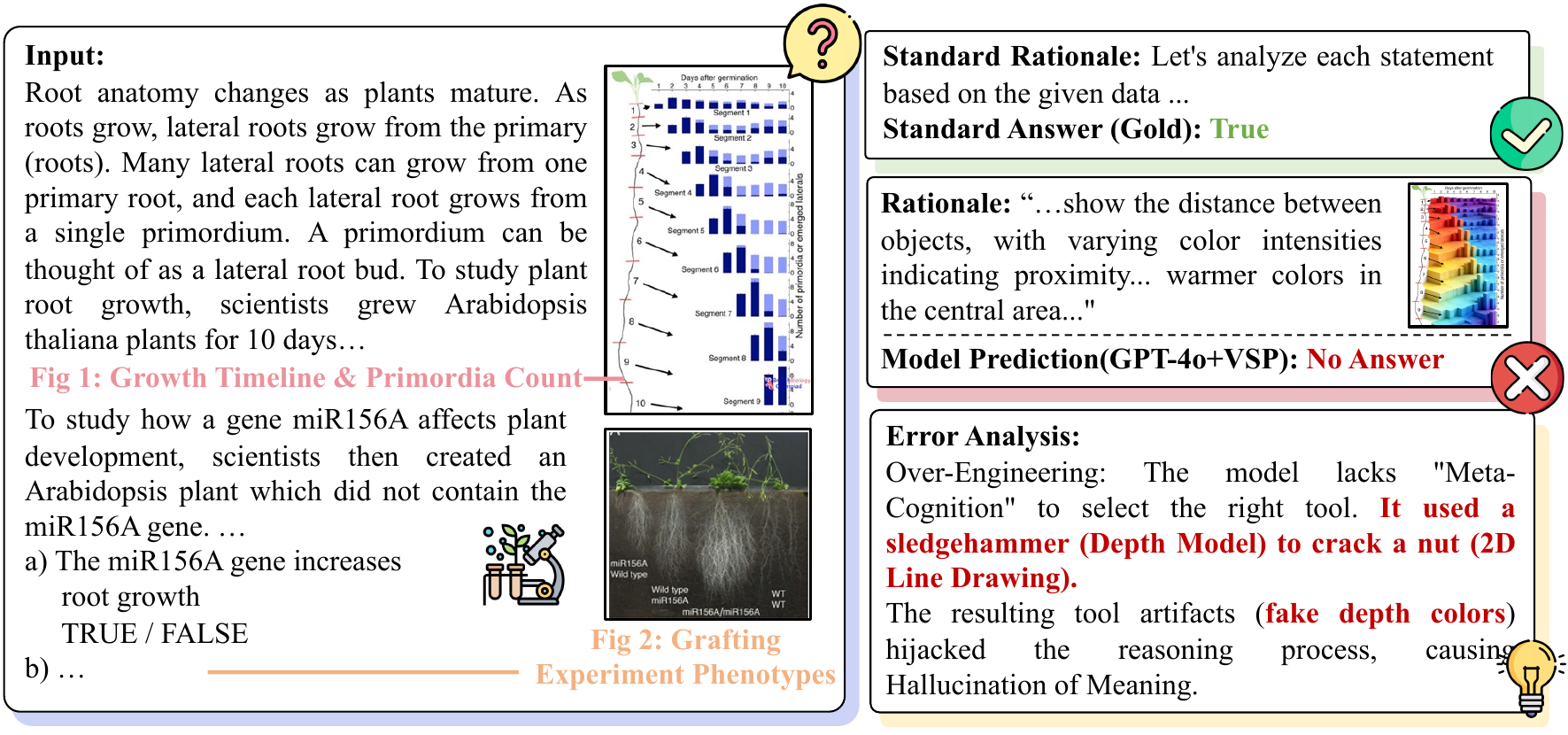} 
    \vspace{-2mm}
    \caption{Analysis of reasoning error examples on GPT-4o+VisualSketchpad (VSP).}
    \label{fig:case-9}
\end{figure}

\begin{figure}[t]
    \centering
    \vspace{-3mm}
    \includegraphics[width=0.92\textwidth]{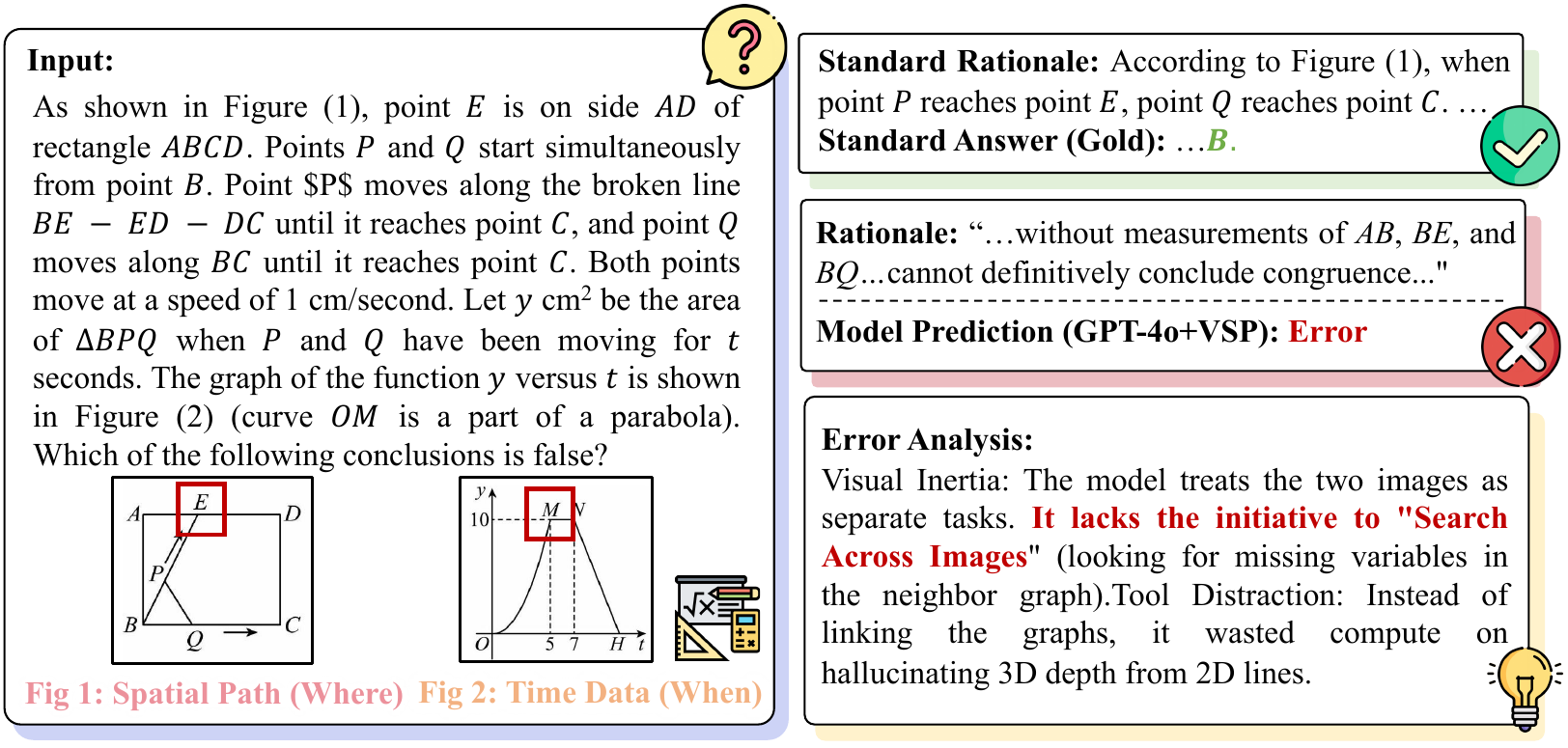} 
    \vspace{-2mm}
    \caption{Analysis of reasoning error examples on GPT-4o+VisualSketchpad (VSP).}
    \label{fig:case-10}
\end{figure}

\vspace{-2mm}\section{Additional Details for ``Thinking with Images''}\vspace{-1mm}
\label{append:think-with-image}
\subsection{Tool-based ``Thinking with Images'' (GPT-4o + VisualSketchpad)}\vspace{-1mm}

Inspired by \citet{cheng2025comt,tong2025thinking}, we instantiate the tool-based ``Think-with-Image'' paradigm by augmenting GPT-4o with VisualSketchpad. This integration enables the model to draw primitives and text, highlight or blur regions, and crop or zoom. We adapt VisualSketchpad’s prompting strategy, originally optimized for single-image tasks, to our multi-image framework.

Qualitatively, the system exhibits distinct limitations on OMIBench: (1) fixating on isolated images while neglecting other crucial information for the task; (2) performing redundant edits that yield no new information; and (3) failing to spatially align objects across images. These behaviors, rare in single-image settings, highlight the limited transferability of existing visual tools to complex multi-image reasoning. More detailed case analyses are provided in Figure~\ref{fig:case-9} \& \ref{fig:case-10}.

\vspace{-2mm}\subsection{Internal ``Thinking with Images'' (EMU-3.5-34B)}\vspace{-1mm}

\begin{figure}
    \centering
    \vspace{-3mm}
    \includegraphics[width=0.92\textwidth]{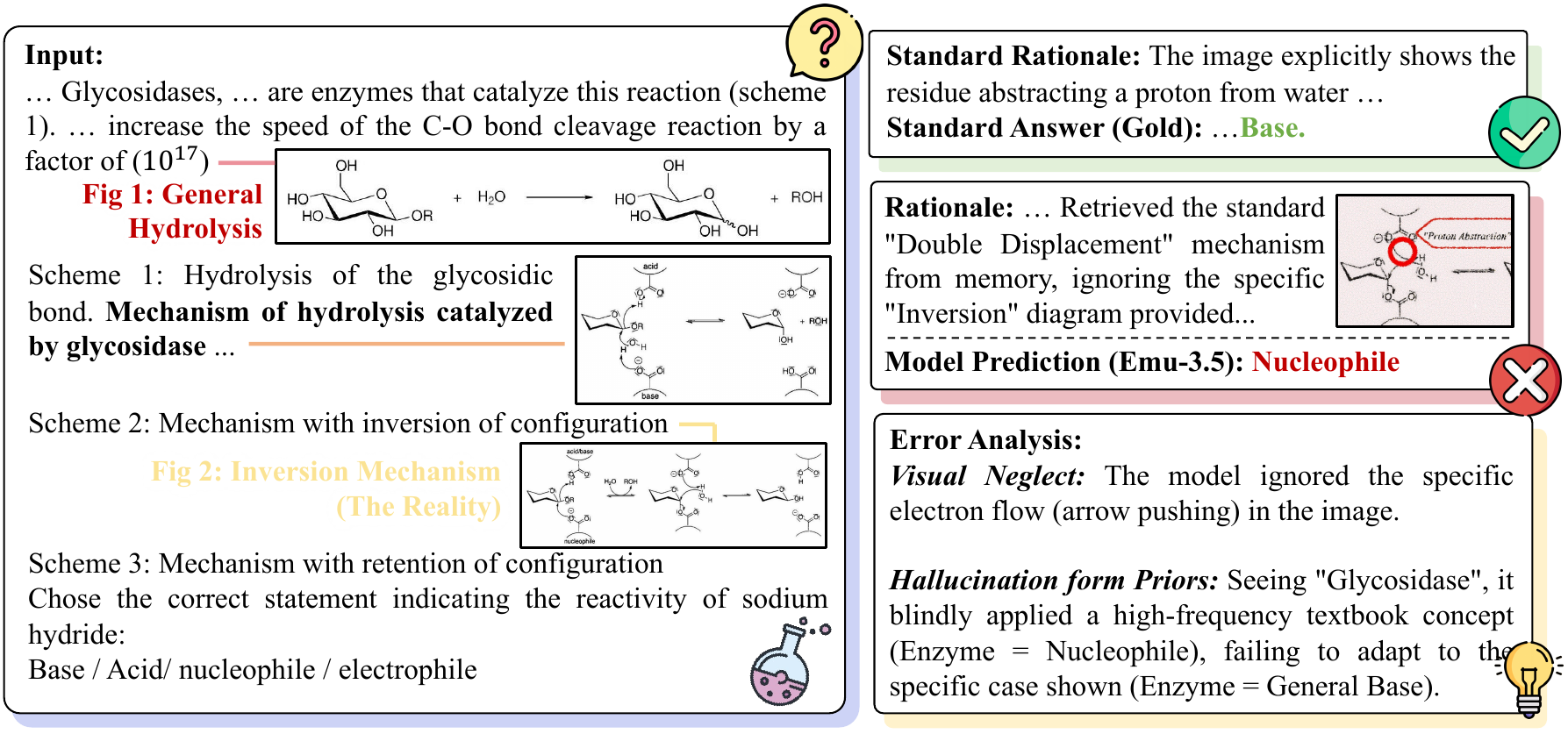} 
    \vspace{-2mm}
    \caption{Analysis of reasoning error examples on Emu-3.5.}
    \label{fig:case-11}
\end{figure}

\begin{figure}
    \centering
    \vspace{-3mm}
    \includegraphics[width=0.92\textwidth]{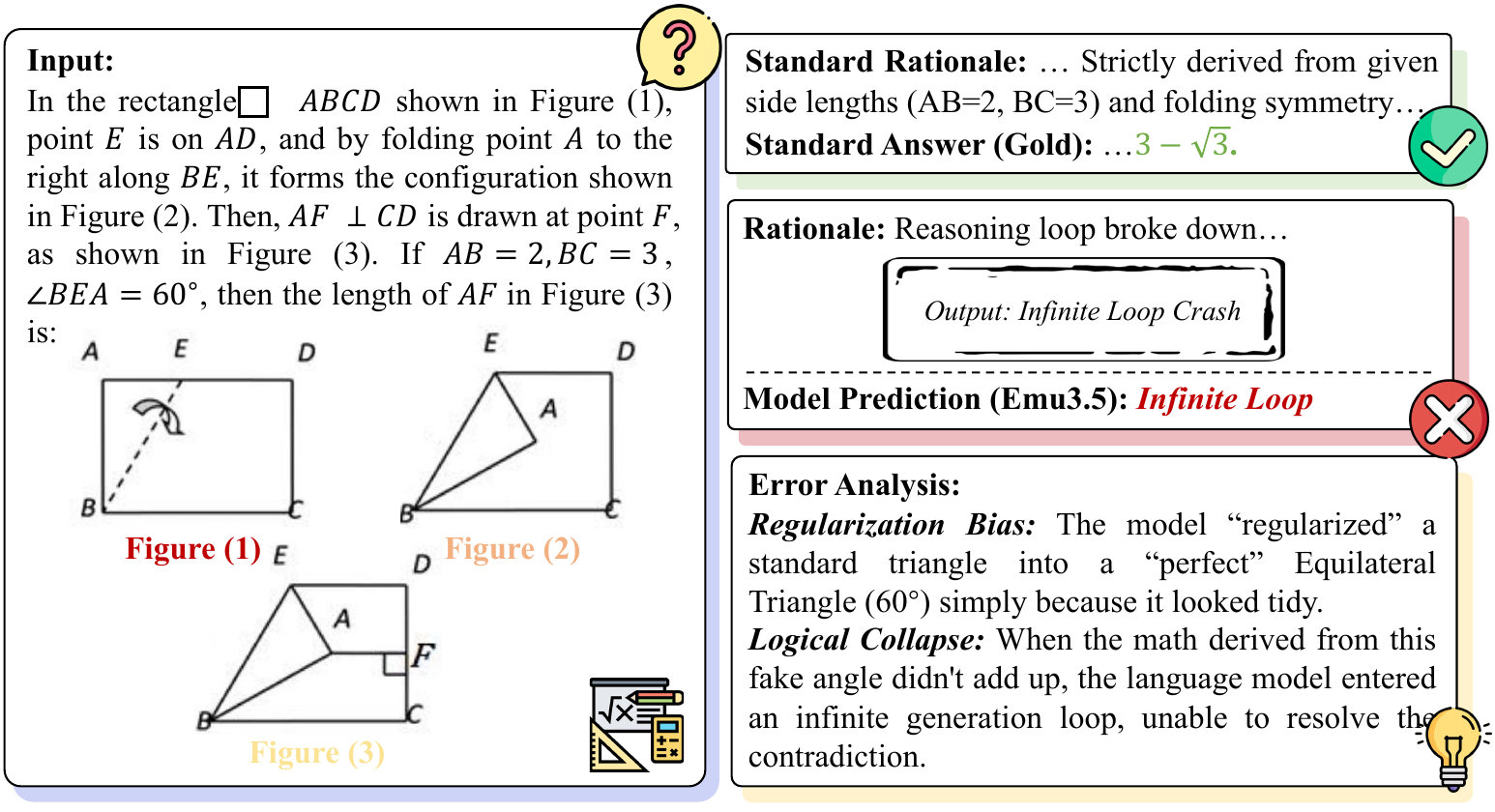} 
    \vspace{-2mm}
    \caption{Analysis of reasoning error examples on Emu-3.5.}
    \label{fig:case-12}
\end{figure}

\begin{figure}
    \centering
    \vspace{-3mm}
    \includegraphics[width=0.92\textwidth]{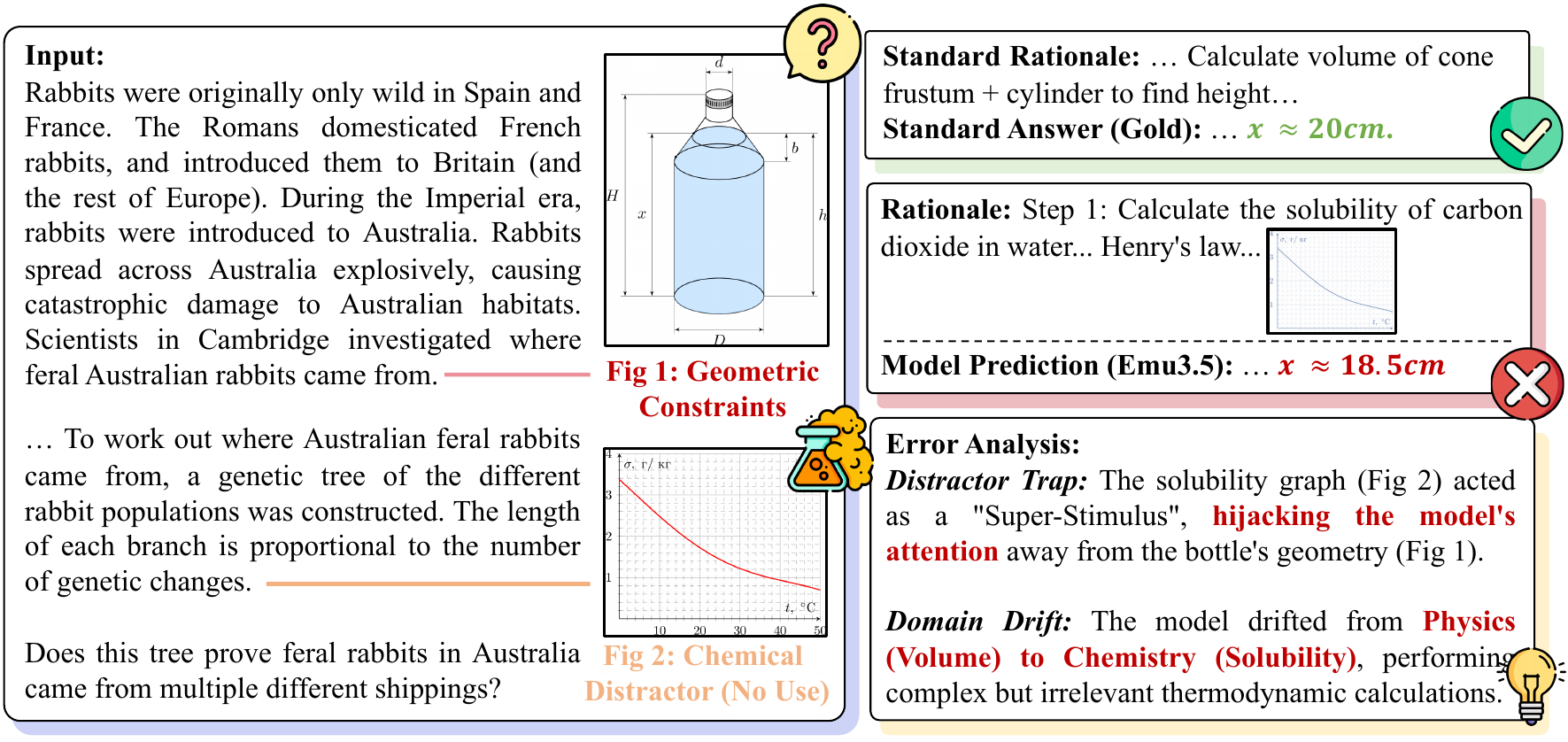} 
    \vspace{-2mm}
    \caption{Analysis of reasoning error examples on Emu-3.5.}
    \label{fig:case-13}
\end{figure}

In the internal ``Think-with-Image'' configuration, we employ EMU-3.5-34B, a unified multimodal generator designed to reason over inputs and synthesize visual content autonomously, eliminating the need for external APIs. This architecture enables the model to perform reasoning that integrates both textual analysis and self-generated visual aids. For inference, we adopt a mixed-precision protocol on 2 NVIDIA A100 80GB GPUs. We set the temperature to 0.3 and top-p to 0.9, a configuration empirically determined to optimally balance diversity and coherence during visual planning. More detailed case analyses are provided in Figure~\ref{fig:case-11}-\ref{fig:case-13}.


\section{Reliability and Stability of GPTScore}
\label{append:gptscore-reliability}

To address concerns about the trustworthiness of model-based evaluation, we report four complementary analyses of GPTScore.

\paragraph{Complementarity with match accuracy.}
Match Accuracy captures exact symbolic agreement, while GPTScore evaluates semantic equivalence under the multimodal context (see Appendix~\ref{append:metrics}). The Spearman correlation between the two metrics over model-level scores is $0.93$, indicating strong agreement and consistency.

\paragraph{Agreement with human ratings and cross-evaluator stability.}
We randomly sample $200$ answers generated by Gemini-3-Pro-Preview and obtain human ratings. We then compare these human scores with GPTScore produced by three different evaluator models. As shown in Table~\ref{tab:gpt-human}, all Spearman correlation coefficients exceed $0.86$ ($p<0.05$), demonstrating both that model-based evaluation is well aligned with human judgment for logic-reasoning-style answers, and that GPTScore is stable across different evaluator models.

\begin{table}[t]
\centering\small
\begin{tabular}{llcccccccccc}
\toprule
\multirow{2}{*}{\textbf{Metric}} & \multirow{2}{*}{\textbf{Evaluator}} & \multicolumn{2}{c}{\textbf{Biology}} & \multicolumn{2}{c}{\textbf{Chemistry}} & \multicolumn{2}{c}{\textbf{Mathematics}} & \multicolumn{2}{c}{\textbf{Physics}} & \multicolumn{2}{c}{\textbf{Total}} \\
\cmidrule(lr){3-4}\cmidrule(lr){5-6}\cmidrule(lr){7-8}\cmidrule(lr){9-10}\cmidrule(lr){11-12}
 &  & Mean & Std & Mean & Std & Mean & Std & Mean & Std & Mean & Std \\
\midrule
GPTScore & Gemini-2.5-flash & 72.49 & 1.01 & 26.61 & 0.95 & 61.57 & 0.88 & 40.33 & 1.17 & 50.98 & 0.42 \\
GPTScore & GPT-5-mini       & 71.49 & 0.33 & 23.85 & 1.02 & 62.85 & 1.25 & 40.21 & 0.99 & 50.80 & 0.45 \\
\midrule
ACC      & Gemini-2.5-flash & 59.74 & 1.29 & 21.08 & 0.69 & 39.71 & 1.19 & 21.76 & 0.76 & 34.66 & 0.29 \\
ACC      & GPT-5-mini       & 60.14 & 1.20 & 21.43 & 0.88 & 36.92 & 0.79 & 24.47 & 0.89 & 34.75 & 0.22 \\
\bottomrule
\end{tabular}
\caption{Stability of GPTScore and Accuracy (\%) across three independent runs (mean and standard deviation).}
\label{tab:stability}
\end{table}

\begin{table}[t]
\centering\small
\begin{tabular}{lcccccccccc}
\toprule
\multirow{2}{*}{\textbf{Setting}} & \multicolumn{2}{c}{\textbf{Biology}} & \multicolumn{2}{c}{\textbf{Chemistry}} & \multicolumn{2}{c}{\textbf{Mathematics}} & \multicolumn{2}{c}{\textbf{Physics}} & \multicolumn{2}{c}{\textbf{Total}} \\
\cmidrule(lr){2-3}\cmidrule(lr){4-5}\cmidrule(lr){6-7}\cmidrule(lr){8-9}\cmidrule(lr){10-11}
 & ACC & Score & ACC & Score & ACC & Score & ACC & Score & ACC & Score \\
\midrule
Original multi-image                & 60.64 & 52.61 & 15.67 & 22.12 & 29.53 & 15.35 & 16.98 & 18.87 & 24.05 & 24.88 \\
Inform.-equivalent single-image & \textbf{64.66} & \textbf{56.63} & \textbf{20.74} & \textbf{26.27} & \textbf{33.49} & \textbf{19.30} & \textbf{22.41} & \textbf{24.53} & \textbf{33.71} & \textbf{29.17} \\
\bottomrule
\end{tabular}
\caption{Performance comparison of GPT-4o between the original multi-image setting and the information-equivalent single-image setting.}
\label{tab:single-vs-multi}
\end{table}

\begin{table}[t]
\centering\small
\setlength{\tabcolsep}{4pt}
\begin{tabular}{p{0.24\textwidth}cc}
\toprule
\textbf{Comparison} & \textbf{Spearman Cor.} & \textbf{$p$-value} \\
\midrule
Human vs.\ GPT-4o-mini      & 0.886 & $<\!0.05$ \\
Human vs.\ GPT-5-mini       & 0.874 & $<\!0.05$ \\
Human vs.\ Gemini-2.5-flash & 0.869 & $<\!0.05$ \\
\bottomrule
\end{tabular}
\caption{Spearman correlation between human ratings and model-based GPTScore.}
\label{tab:gpt-human}
\end{table}

\paragraph{Sampling stability and statistical significance.}
We report the mean and standard deviation of both GPTScore and ACC across three independent sampling runs of Gemini-3-Pro-Preview, evaluated by Gemini-2.5-flash and GPT-5-mini. As shown in Table~\ref{tab:stability}, the standard deviations remain small: for GPTScore, all subject-level deviations are below $1.3\%$ and total-level Std is below $0.5\%$; for ACC, fluctuations are at most $\sim\!1.3\%$ and total-level Std is below $0.3\%$. These variations are much smaller than the effective gains reported in our paper for techniques such as long-CoT thinking, sequential scaling, and in-context learning, which exceed $5\%$ and in most cases exceed $10\%$. Moreover, paired comparisons for Qwen-VL-32B-Thinking vs.\ Instruct, Qwen-VL-32B-Instruct sequential scaling (1 vs.\ 16 samples), and Qwen-VL-32B-Instruct textual ICL (0-shot vs.\ 3-shot) over five repeated runs yield $p$-values all below $0.05$, indicating that our conclusions are statistically stable.

\begin{table}[t]
\centering
\begin{tabular}{lc}
\toprule
\textbf{Evaluation subject} & \textbf{Accuracy} \\
\midrule
Human Expert A (avg.)       & 82.69\% \\
Human Expert B (avg.)       & 80.77\% \\
Human Non-expert A (avg.)   & 57.69\% \\
Human Non-expert B (avg.)   & 61.53\% \\
Gemini-3-Pro (best current) & 48.08\% \\
\bottomrule
\end{tabular}
\caption{Performance gaps between human experts/non-experts and the current strongest model on a 52-problem subset of OMIBench.}
\label{tab:human-baseline}
\end{table}

\paragraph{Partial-credit evaluation (future work).}
Although GPTScore already provides graded semantic judgments under contextual constraints, we acknowledge that purely binary or near-binary judgments may underweight creative or partially correct solutions on highly open-ended OMIBench items. As a future extension, we plan to incorporate a rubric-guided partial-scoring mechanism following recent practice in scientific-reasoning evaluation~\citep{openai2025frontierscience}, and to release the corresponding rubrics and evaluator prompts together with the benchmark.

\begin{table}[t]
\centering\small
\begin{tabular}{lcccccccccc}
\toprule
\multirow{2}{*}{\textbf{Model}} & \multicolumn{2}{c}{\textbf{Biology}} & \multicolumn{2}{c}{\textbf{Chemistry}} & \multicolumn{2}{c}{\textbf{Mathematics}} & \multicolumn{2}{c}{\textbf{Physics}} & \multicolumn{2}{c}{\textbf{Total}} \\
\cmidrule(lr){2-3}\cmidrule(lr){4-5}\cmidrule(lr){6-7}\cmidrule(lr){8-9}\cmidrule(lr){10-11}
 & ACC & Score & ACC & Score & ACC & Score & ACC & Score & ACC & Score \\
\midrule
InternVL3.5-8B       & 47.39 & 36.95 & 17.51 & 17.97 & 27.91 & 32.33 & 17.69 & 17.69 & 26.59 & 26.14 \\
\;+ CMMCoT           & \textbf{48.19} & 36.55 & \textbf{17.97} & \textbf{19.35} & \textbf{29.77} & \textbf{35.58} & \textbf{19.81} & \textbf{18.63} & \textbf{28.11} & \textbf{27.65} \\
\;+ MMDU             & 46.18 & 36.55 & 17.51 & 16.59 & 26.98 & 31.86 & 16.75 & 16.51 & 25.76 & 25.30 \\
\midrule
Qwen3-VL-8B-Instruct & 46.59 & 43.37 & 16.13 & 16.59 & 27.21 & 29.07 & \textbf{20.05} & \textbf{18.40} & 26.74 & 26.29 \\
\;+ CMMCoT           & \textbf{49.00} & \textbf{49.40} & \textbf{18.43} & \textbf{18.43} & \textbf{31.63} & \textbf{30.70} & 19.58 & 17.92 & \textbf{28.86} & \textbf{28.11} \\
\;+ MMDU             & 45.78 & 41.77 & 16.13 & 15.67 & 25.81 & 29.53 & 19.34 & 18.16 & 25.91 & 25.91 \\
\bottomrule
\end{tabular}
\caption{SFT results on OMIBench with different multi-image instruction-tuning datasets.}
\label{tab:sft}
\end{table}
\begin{table}[t]
\centering\small
\begin{tabular}{lcccccccccc}
\toprule
\multirow{2}{*}{\textbf{Method}} & \multicolumn{2}{c}{\textbf{Biology}} & \multicolumn{2}{c}{\textbf{Chemistry}} & \multicolumn{2}{c}{\textbf{Mathematics}} & \multicolumn{2}{c}{\textbf{Physics}} & \multicolumn{2}{c}{\textbf{Total}} \\
\cmidrule(lr){2-3}\cmidrule(lr){4-5}\cmidrule(lr){6-7}\cmidrule(lr){8-9}\cmidrule(lr){10-11}
 & ACC & Score & ACC & Score & ACC & Score & ACC & Score & ACC & Score \\
\midrule
GPT-4o                & \textbf{60.64} & \textbf{52.61} & 15.67 & \textbf{22.12} & 29.53 & 15.35 & 16.98 & 18.87 & 24.05 & 24.88 \\
\;+ SlowPerception    & 51.41 & 52.21 & 7.83  & 17.97 & 18.14 & 6.98  & 1.89  & 14.39 & 17.50 & 19.70 \\
\;+ CogFlow           & 47.39 & 40.56 & 7.37  & 13.36 & 8.84  & 4.42  & 6.37  & 2.36  & 15.08 & 12.05 \\
\;+ Visual Sketchpad  & 18.88 & 17.67 & \textbf{16.13} & 17.97 & 9.53  & 13.72 & 15.33 & 16.75 & 14.24 & 16.14 \\
\midrule
GPT-5                 & 68.13 & \textbf{62.55} & \textbf{23.96} & 29.03 & 39.30 & 56.51 & 20.52 & 40.80 & 36.23 & 48.11 \\
\;+ SlowPerception    & 69.88 & \textbf{63.45} & 23.04 & 29.95 & 39.07 & 56.98 & \textbf{20.99} & \textbf{42.22} & 36.44 & 49.02 \\
\;+ CogFlow           & 72.29 & 62.25 & 23.50 & \textbf{33.64} & 38.84 & \textbf{57.44} & 20.52 & 41.04 & 36.74 & \textbf{49.17} \\
\;+ Visual Sketchpad  & \textbf{73.90} & 61.85 & 22.12 & 30.88 & \textbf{39.53} & 56.51 & 20.05 & 38.21 & \textbf{36.89} & 47.42 \\
\bottomrule
\end{tabular}
\caption{Results of different external-tool integration frameworks on OMIBench.}
\label{tab:tools}
\end{table}

\section{Human Baseline Details}
\label{append:human-baseline}

To calibrate the difficulty of OMIBench against human performance, we conduct a small-scale human study on a uniformly sampled 52-problem subset spanning biology, chemistry, mathematics, and physics. We recruit four participants: two domain experts (PhD/Master's students in STEM-related fields) and two trained non-experts (undergraduates without direct domain specialization). None of the participants has seen OMIBench before the study.

Before formal evaluation, all participants complete 10 familiarization problems that are not included in the reported subset. This stage is used only to acquaint them with the OMIBench input format, which often requires coordinating information across multiple images, textual problem statements, and sometimes auxiliary answer choices. During the actual evaluation, participants are allowed to inspect all provided images and text for each problem and are asked to provide a final answer in the same problem setting as the benchmark.

We report average accuracy for each participant group in Table~\ref{tab:human-baseline}, together with the strongest current model on the same subset for reference. The results show that experts remain substantially above current LVLMs, while even trained non-experts still outperform the best model by a clear margin. This gap indicates that OMIBench is difficult but still solvable for humans with sufficient scientific background, making it a meaningful benchmark for measuring progress in multi-image Olympiad-level reasoning.

\section{Single-Image Control Details}
\label{append:single-vs-multi}

To distinguish the effect of cross-image reasoning from confounders such as visual volume, input length, and OCR noise, we construct an information-equivalent single-image control. For each problem, all images are concatenated in their original logical order into a single composite image, while the question text and answer choices remain unchanged.

\section{Detailed Results for Training and External Tools}
\label{append:sft-tools}

This appendix provides the full numerical results referenced in Section~\ref{sec:sft-tools}.

\paragraph{SFT on multi-image instruction-tuning data.}
Table~\ref{tab:sft} reports the per-subject ACC and Score for InternVL3.5-8B and Qwen3-VL-8B-Instruct fine-tuned on CMMCoT~\citep{cmmcot2026} and MMDU~\citep{mmdu2024}. CMMCoT consistently yields modest improvements on both backbones (Total ACC $26.59\!\to\!28.11$ and $26.74\!\to\!28.86$ respectively), while MMDU yields no improvement or slight degradation, suggesting that simple multi-image understanding data is insufficient for Olympiad-level reasoning.

\paragraph{External-tool integration.}
Table~\ref{tab:tools} reports the per-subject ACC and Score of three external-tool integration frameworks on top of GPT-4o and GPT-5. GPT-4o consistently degrades when augmented with any of the three tools, while GPT-5 shows modest, inconsistent gains.

\end{document}